\documentclass[letterpaper, 10 pt, conference]{ieeeconf} 
\usepackage[utf8]{inputenc}

\IEEEoverridecommandlockouts
\usepackage{units}
\usepackage{algorithm, algorithmic}
\usepackage{amsmath}
\usepackage{amssymb}
\usepackage{tabularx}
\usepackage{booktabs}
\usepackage{cite} 
\usepackage{graphicx}
\usepackage{subcaption}
\usepackage{mwe}
\usepackage{hyperref}
\usepackage[svgnames]{xcolor}
\usepackage{multirow}
\usepackage{makecell}
\usepackage{xcolor,colortbl}
\hypersetup{
 colorlinks = true,
 linkcolor=black,
 urlcolor=Blue}

\usepackage{amsthm} 

\usepackage[colorinlistoftodos]{todonotes}

\newif\ifcorrectingmode
\correctingmodefalse
\usepackage[normalem]{ulem}
\ifcorrectingmode
	
	\newcommand{\deleted}[1]{\textcolor{Black}{\ifmmode\text{\sout{\ensuremath{#1}}}\else\sout{#1}\fi}}
	\newcommand{\deletedequation}[2]{\textcolor{Black}{\centerline{Removed equation (#1)}}}
\else
	
	\newcommand{\deleted}[1]{}
	\newcommand{\deletedequation}[2]{}
\fi

\usepackage[printonlyused]{acronym}
\newacro{DOF}{degree of freedom}
\newacroplural{DOF}{degrees of freedom}
\acrodefplural{DOF}[DOFs]{degrees of freedom}
\newacro{iLQR}{Iterative Linear-Quadratic Regulator}
\newacro{CT}{Control Toolbox}
\newacro{EOM}{equations of motion}
\newacro{OC}{Optimal Control}
\newacro{LQR}{linear-quadratic regulator}
\newacro{PD}{proportional derivative}
\newacro{MPC}{Model Predictive Control}
\newacro{LQ}{linear quadratic}
\newacro{LQOC}{Linear-Quadratic Optimal Control}
\newacro{TO}{Trajectory Optimization}
\newacro{DDP}{Differential Dynamic Programming}
\newacro{COM}{center of mass}
\newacro{COP}{center of pressure}
\newacro{NLP}{nonlinear program}
\newacro{MLP}{Multilayer Perceptron}
\newacro{SLQ}{Sequential Linear-Quadratic}
\newacro{HAA}{hip abduction adduction}
\newacro{AD}{automatic differentiation}
\newacro{HJB}{Hamilton–Jacobi–Bellman}
\newacro{BC}{Behavioral Cloning}
\newacro{IRL}{Inverse Reinforcement Learning}
\newacro{IL}{Imitation Learning}
\newacro{RL}{Reinforcement Learning}
\newacro{MEN}{mixture-of-experts network}
\newacro{MILE}{mixture of implicitly localized experts}
\newacro{MELE}{mixture of explicitly localized experts}
\newacro{DL}{Deep Learning}
\newacro{WBC}{whole-body controller}
\newacro{MDP}{Markov Decision Process}
\newacro{RNN}{Recurrent neural network}
\newacro{PPO}{Proximal Policy Optimization}

\def\TheTitle{Learning Arm-Assisted Fall Damage Reduction and Recovery for Legged Mobile Manipulators}

\title{\LARGE \bf \TheTitle} 
\author{Yuntao Ma, Farbod Farshidian, Marco Hutter
    \thanks{This work was supported by the Max Planck ETH Center for Learning Systems, the Swiss National Science Foundation (SNSF) through project 166232, 188596, the National Centre of Competence in Research Robotics (NCCR Robotics), and the European Union's Horizon 2020 (grant agreement No.852044 and No.101016970). Moreover, this work has been conducted as part of ANYmal Research, a community to advance legged robotics.}%
    \thanks{All authors are with the Robotic Systems Lab, ETH Z\"u{}rich, Switzerland. Email: {\tt\footnotesize mayun@ethz.ch}}%
}

\begin{document}

\maketitle
%
\thispagestyle{empty}
\pagestyle{empty}
%
%
\begin{abstract}

Adaptive falling and recovery skills greatly extend the applicability of robot deployments. In the case of legged mobile manipulators, the robot arm could adaptively stop the fall and assist the recovery. Prior works on falling and recovery strategies for legged mobile manipulators usually rely on assumptions such as inelastic collisions and falling in defined directions to enable real-time computation. This paper presents a learning-based approach to reducing fall damage and recovery. An asymmetric actor-critic training structure is used to train a time-invariant policy with time-varying reward functions. In simulated experiments, the policy recovers from 98.9\% of initial falling configurations. It reduces base contact impulse, peak joint internal forces, and base acceleration during the fall compared to the baseline methods. The trained control policy is deployed and extensively tested on the ALMA robot hardware. A video summarizing the proposed method and the hardware tests is available at \href{https://youtu.be/avwg2HqGi8s}{https://youtu.be/avwg2HqGi8s}.

\end{abstract}

%

%
%


\section{Introduction}\label{sec:introduction}

Legged mobile manipulators have a high potential for practical applications because of their hardware capability for manipulation and traversing non-flat terrains. For such applications, specialized payloads such as sensors and end-effectors may be mounted on the robot. One factor limiting such robots' deployment is that, in case of falling, the payload and the manipulator are easily damaged, making these robots too fragile for field deployment. To this end, damage reduction during the fall and recovery from failure are considered among the key remaining challenges in the legged robotic field \cite{christensen2021roadmap}.
Both falling and recovery are contact-rich maneuvers that require the robot to make meaningful interactions with the ground. The frequent contact switches pose a challenge for the control methods, but there has recently been tremendous progress {\color{Black}in} these tasks.

To reduce the fall damage, robot controllers may plan fixed~\cite{yun2014tripod} or adaptive~\cite{ha2015multiple} contact sequences to reduce the peak acceleration, contact forces on the robot bodies, or other damage criteria. The robot may execute ukemi-like motions~\cite{ogata2007falling} after making contact with the ground to soften the fall. However, these methods often rely on restrictive assumptions such as inelastic and non-sliding collisions or simplifications such as falling in the sagittal or frontal plane. Furthermore, the limbs are sometimes assumed to be sufficiently agile to track the adaptive contact sequence plans\cite{kumar2017learning}, while this assumption does not hold for robots with heavier legs and more restrictive joint velocity limits. 

For fall recovery, planning-based methods often rely on accurate estimation of state and contact points between the robot and the environment, which poses additional challenges due to uncertainties in contact estimation and variations in the initial contact configuration. A recent and promising solution for the legged robot's fall recovery is to use reinforcement learning (RL)~\cite{hwangbo2019learning, lee2019robust, zhang2022accessibility}, where many of the heuristics and simplifications are avoided.

\begin{figure}
    \centering
    \subfloat[]{\includegraphics[width=0.15\textwidth]{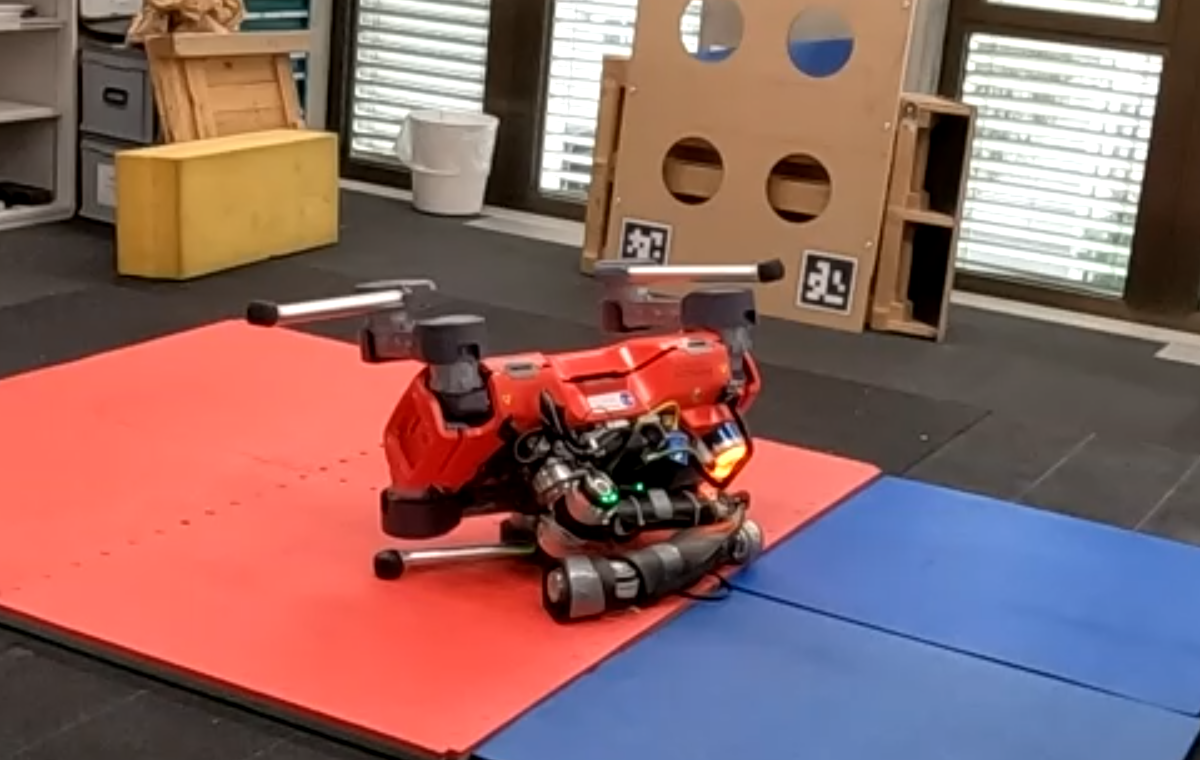}
    \vspace{-5pt}
    \label{subfig:seq0}}
    \hfil
    \subfloat[]{\includegraphics[width=0.15\textwidth]{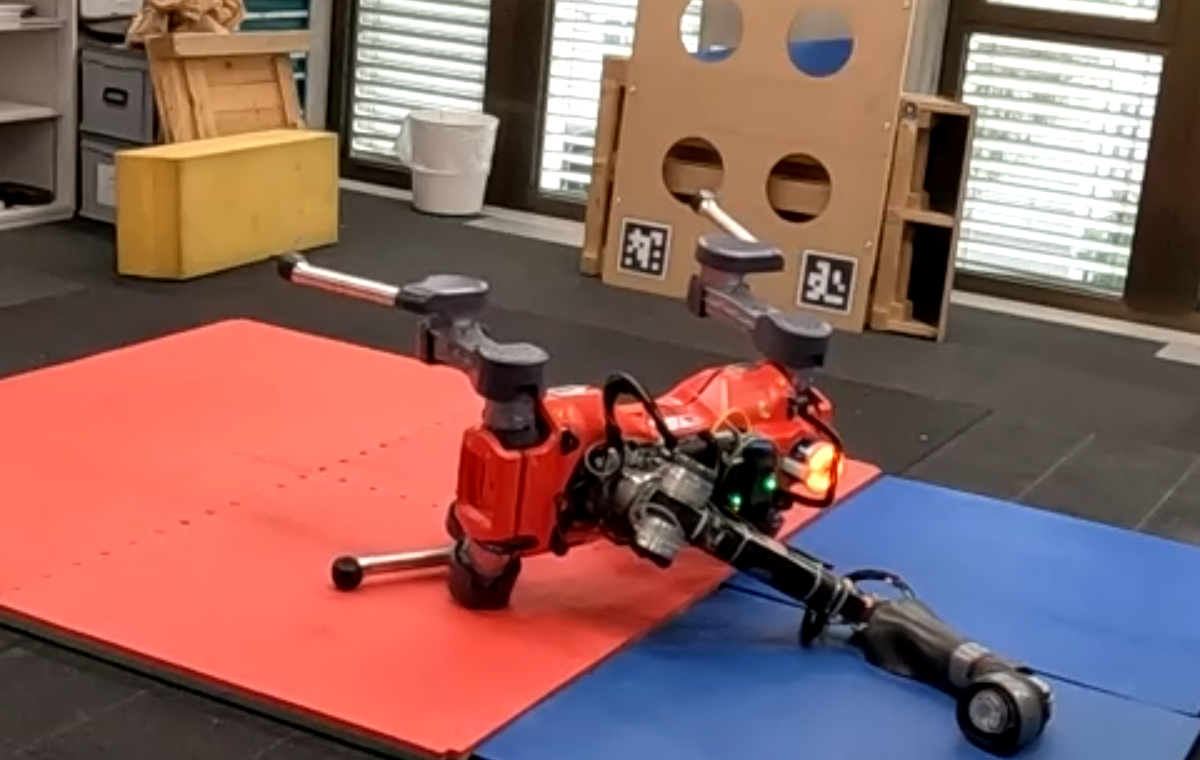}
    \vspace{-5pt}
    \label{subfig:seq1}}
    \hfil
    \subfloat[]{\includegraphics[width=0.15\textwidth]{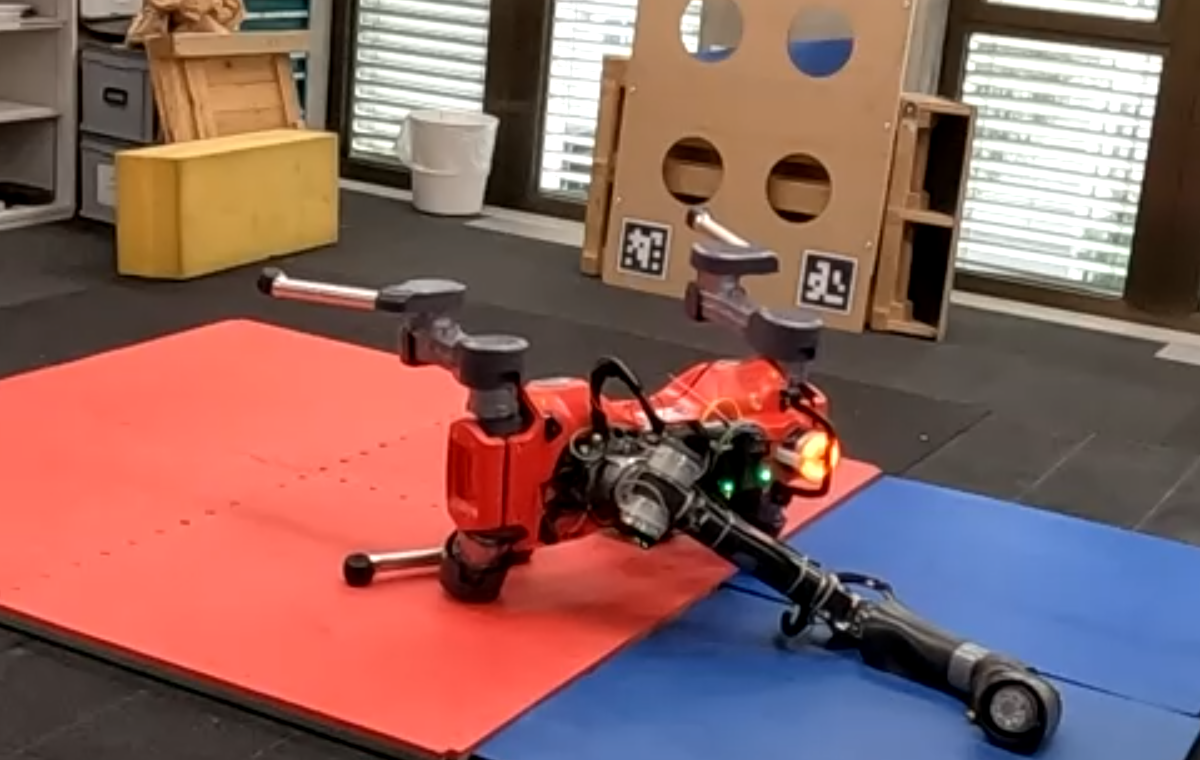}
    \vspace{-5pt}
    \label{subfig:seq2}}
    \hfil
    \subfloat[]{\includegraphics[width=0.15\textwidth]{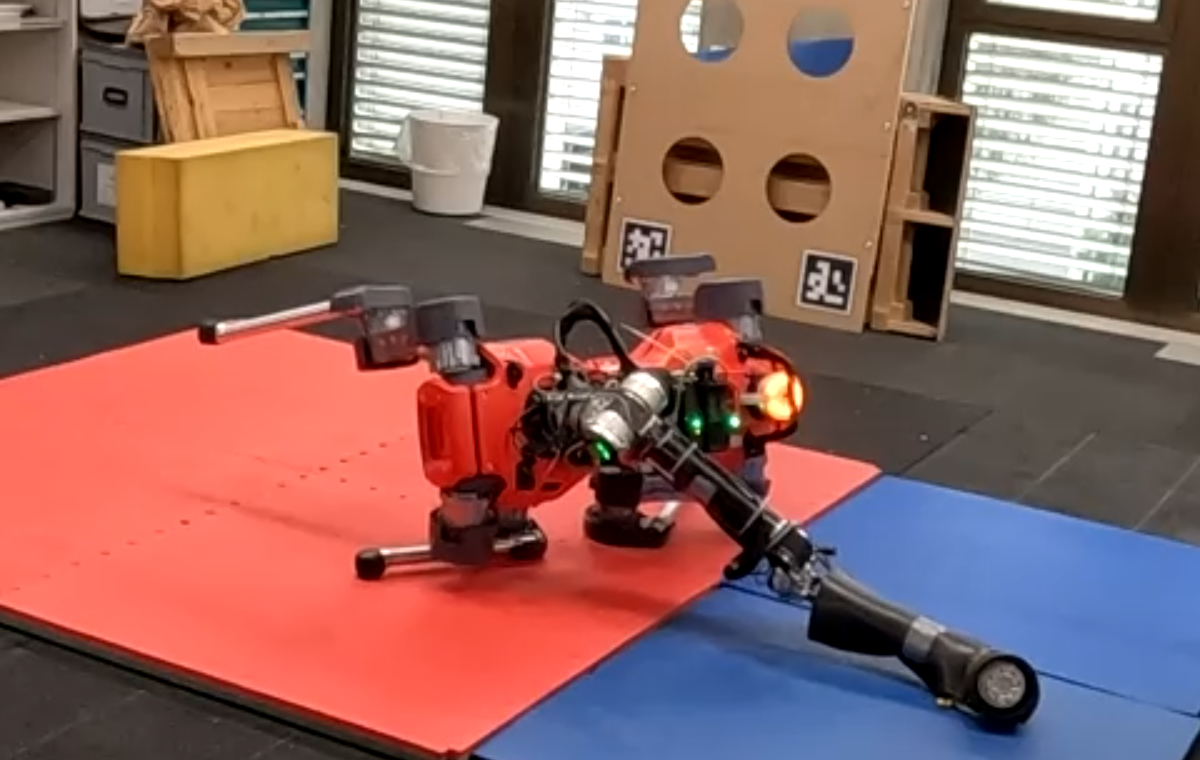}
    \vspace{-5pt}
    \label{subfig:seq3}}
    \hfil
    \subfloat[]{\includegraphics[width=0.15\textwidth]{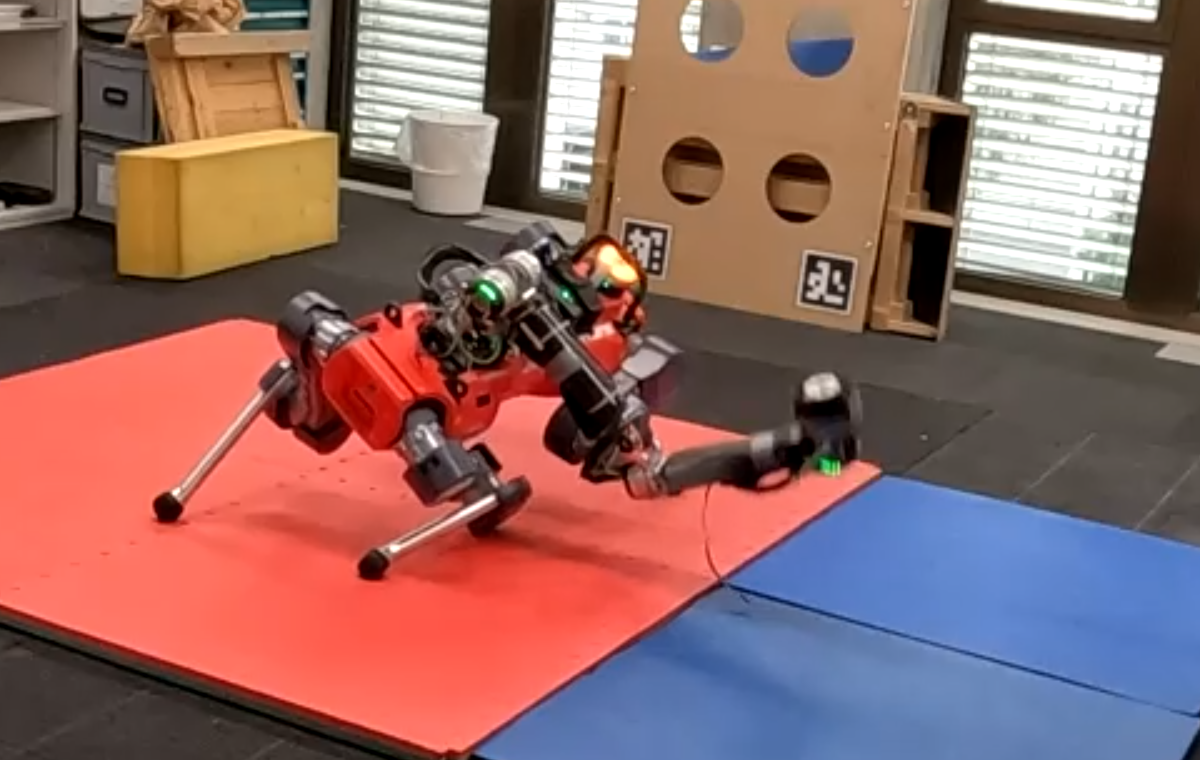}
    \vspace{-5pt}
    \label{subfig:seq4}}
    \hfil
    \subfloat[]{\includegraphics[width=0.15\textwidth]{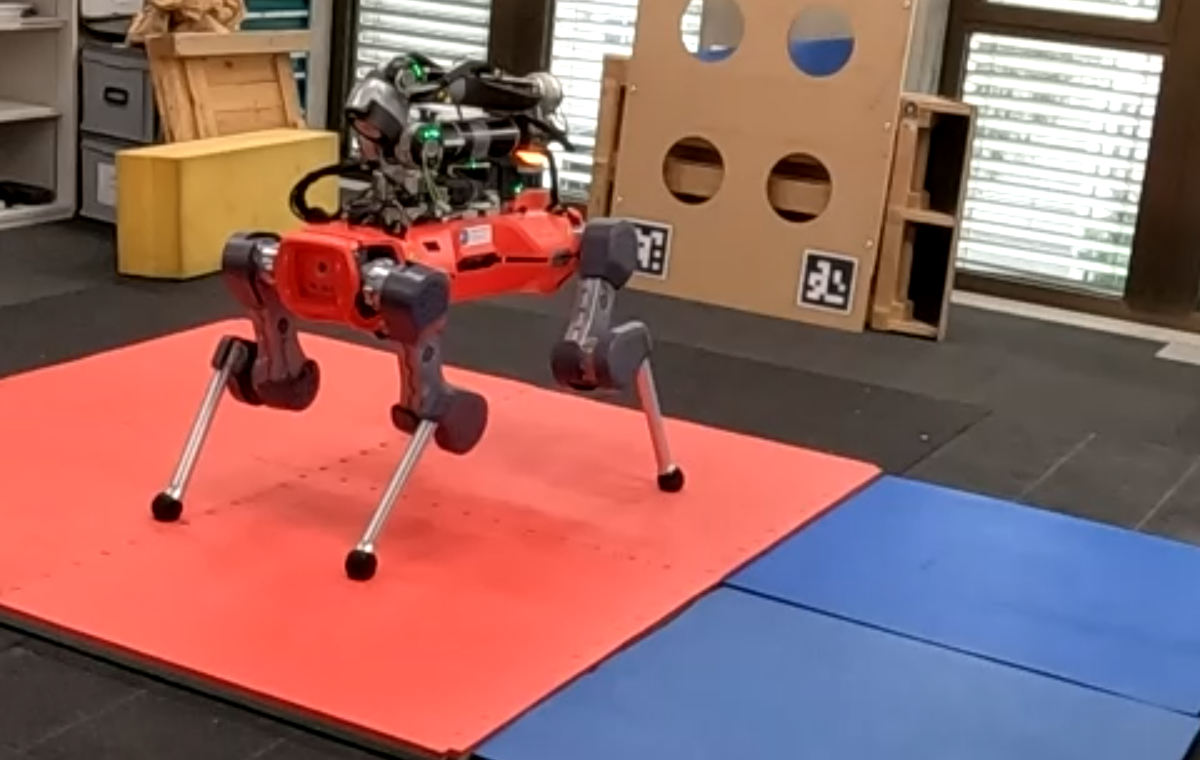}
    \vspace{-5pt}
    \label{subfig:seq5}}
    \vspace{2pt}
    \caption{ALMA robot recovering from a fall.}
    \vspace{-6pt}
    \label{fig:title}
\end{figure}

This paper explores a generic learning-based approach to reducing fall damage and recovering falls for legged manipulators with minimal simplifications. We extend our previous learning-based recovery methods \cite{hwangbo2019learning, lee2019robust} by using the arm to reduce fall damage and assist in fall recovery. To this end, we train a single failing-recovery policy for legged mobile manipulators with an on-policy asymmetric actor-critic method with time-{\color{Black}varying} rewards and avoid brittle and laborious tuning of the relative weight between the motion smoothness and the time to complete the recovery. The main contributions are summarized as follows:

\begin{itemize}
    \item A quantitative comparison between the policy and currently deployed emergency controllers (freezing and damping the drives) that shows reductions in peak instantaneous impulse on the base, peak joint internal force, and base acceleration during the fall. Ablation studies also show an improvement in recovery behavior compared to the symmetric and time-variant version of the policy.
    \item Simulated results that show the policy being capable of adaptively stopping the falling with the arm and performing arm-assisted recovery to resume the stance configuration within a specified time horizon. We also show hardware validations of the presented adaptive behaviors on the ALMA~\cite{bellicoso2019alma} hardware highlighted in Fig.\ref{fig:title} in both falling and recovery scenarios.
    \item Adaptation of our proposed method to other tasks, namely bringing the robot to the resting configuration and simple self-righting from the fall. Hardware validations of these control policies are also included. 
\end{itemize}

%
%
\section{Related Work}


\subsection{Fall damage reduction}\label{ssec:background_fall}

Safe{\color{Black}-}falling is intensively studied in the context of humanoid robots. It is usually formulated as an optimization problem to minimize momentum change~\cite{ha2015multiple, goswami2014direction, wang2017real}, impact force~\cite{yun2014tripod}, acceleration~\cite{kumar2017learning} or a combination of damage criteria~\cite{fujiwara2006towards,samy2017qp}.

Traditionally, a defined contact sequence is specified to reduce the planning variables. Ogata et al.~\cite{ogata2007falling} proposed implementing ukemi motion when falling forward, and Yun et al.~\cite{yun2014tripod} proposed tripod fall to reach early contact with the ground to reduce transferring of potential energy to kinetic energy. For the specified contact sequence, the contact location could be computed directly~\cite{ogata2007falling} or selected with an RL policy~\cite{yun2014tripod}. 

More recently, methods that allow for adaptive contact sequences were proposed. Ha and Liu~\cite{ha2015multiple} defined a {\color{Black}Markov Decision Process (MDP)} with a pre-defined contact sequence and trained a policy to minimize the vertical instantaneous impulse with the presented contact graphs. Kumar et al.~\cite{kumar2017learning} used RL with a mixture-of-expert structure, where each expert corresponds to a potential contact body and selects the next contact body in the contact sequence during the fall{\color{Black},} based on each expert's value estimation. 

These methods for planning multiple contacts restrict the falling to the sagittal plane to reduce the computation cost. This simplification was relaxed in~\cite{goswami2014direction} and~\cite{samy2017qp}, where the latter solves a multi-objective Quadratic Programming (QP) for post-impact to make the controller act as active compliance to reduce the damage. {\color{Black}However}, there are still other restrictive assumptions, such as inelastic contact when the robot falls.

For small humanoids, a fully-stretched arm configuration is proposed to stop falling quickly~\cite{yun2014tripod,goswami2014direction}. However, such singularity configurations are undesirable for heavy robots such as ALMA ({\color{Black}{approximately}} \unit[58]{kg}) because of the high stress it induces on the drives during the impacts.

\subsection{Fall recovery}\label{sec:background_recover}

Similar to falling, {\color{Black}recovery controllers impose many simplifications and heuristics to reduce the computational cost}. {\color{Black}Moreover}, legged robots may have unreliable state estimation and uncertain contacts after falling{\color{Black}, which} pose additional challenges to optimization-based control methods. So far, most optimization-based fall recovery of legged robots usually require heuristics such as pre-defined control sequences~\cite{stuckler2006getting}, specified state transitions~\cite{castano2019design}, or model simplifications~\cite{saranli2004model}. These heuristics are typically designed for specific robots and require extra engineering effort to transfer onto other platforms.

On the other hand, learning-based control methods are promising alternatives because they rely less on heuristics and simplifications. Hwangbo et al.~\cite{hwangbo2019learning} and Lee et al.~\cite{lee2019robust} were among the first to explore training RL-based control policies for the fall recovery of quadrupedal robots. They formulated the self-righting problem as an infinite MDP, rewarding the agent for restoring to the default resting joint position and upright orientation. A behavior selector was subsequently trained from a Finite State Machine (FSM) to hierarchically combine self-righting, standing, and locomotion for robust control of the ANYmal robot. More recently, Zhang et al.~\cite{zhang2022accessibility} proposed the K-Access algorithm for state-space clustering to improve the sample efficiency of the learning process.
One remaining challenge for the RL-based fall recovery controller to this day is the arduous reward-tuning to find the right balance between the behavioral rewards and the task rewards to produce recovery behavior while avoiding jerky motions. {\color{Black}In this work, we tackle this issue by using time-varying reward functions to reduce motion constraints before the desired recovery time.}

\section{Method}


\begin{figure}
    \centering
    \includegraphics[width=0.48\textwidth]{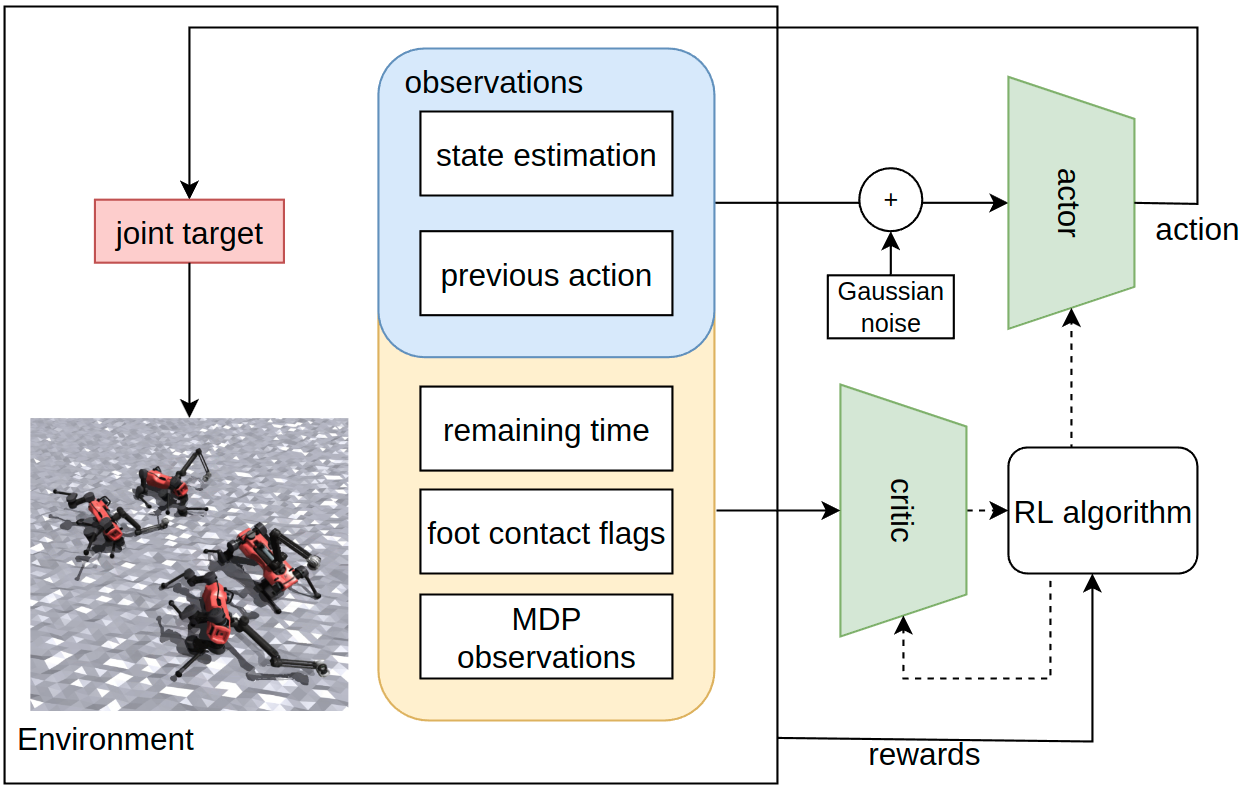}
    \vspace{5pt}
    \caption{Overview of the training pipeline. The actor observes minimal observation required for completing the fall recovery, and the critic has access to privileged information that improves the value function estimation. The actor's output is converted to the joint position targets for the robot's joint PD controller.}
    \label{fig:reward_timing}
\end{figure}

Fig.~\ref{fig:reward_timing} summarises our training pipeline. The control policy is trained with RL to reduce the fall damage from randomized initial falling configurations and recover within a fixed time. We formulate this task as a finite{\color{Black}-}horizon MDP where the robot receives a reward at the end of each episode based on minimizing damage criteria and recovering to the stance posture. This work employs an asymmetric actor-critic~\cite{pinto2017asymmetric} training setup, where the critic receives noiseless and privileged observations. In the following, we outline the main components of our learning pipeline.


\subsection{State initialization and rollout}

\begin{figure}[t]
    \centering
    \subfloat[]{\includegraphics[width=0.11\textwidth]{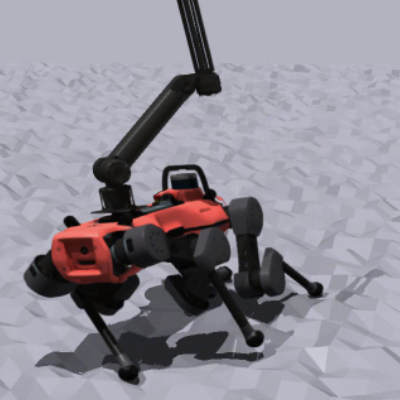}
    \vspace{-5pt}
    \label{subfig:tripping1}}
    \hfil
    \subfloat[]{\includegraphics[width=0.11\textwidth]{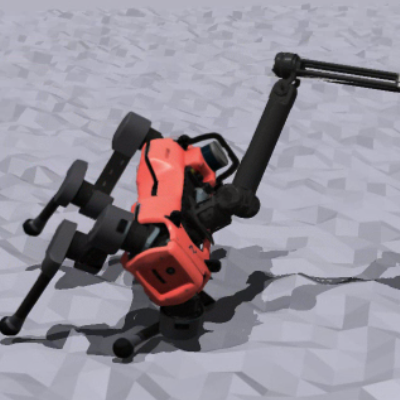}
    \vspace{-5pt}
    \label{subfig:tripping3}}
    \hfil
    \subfloat[]{\includegraphics[width=0.11\textwidth]{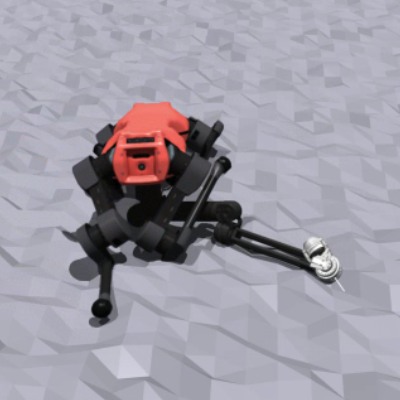}
    \vspace{-5pt}
    \label{subfig:tripping5}}
    \hfil
    \subfloat[]{\includegraphics[width=0.11\textwidth]{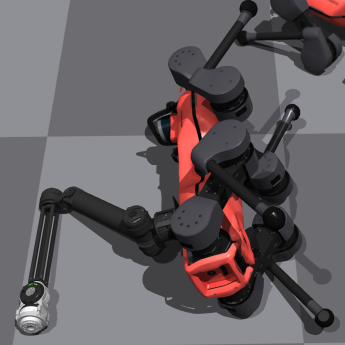}
    \vspace{-5pt}
    \label{subfig:default_joint_pos}}
    \vspace{2pt}
    \caption{\ref{subfig:tripping1}-\ref{subfig:tripping5} Possible initial conditions for the policy training. Our pipeline expects the task controllers to detect and report the fall, and the policy is trained to reduce the damage after switching the controller. \ref{subfig:default_joint_pos} The default joint configuration for the policy training.}
    \vspace{-1pt}
    \label{fig:init}
\end{figure}

Legged mobile manipulators may have different controllers for locomotion, {\color{Black}object manipulation, or dynamic throwing} tasks. As the state distribution (i.e., leg contacts, base orientation, etc.) of the robot may largely vary when performing different tasks, it is hard to train a fall detector that suits all controllers. Therefore, we expect the task controllers to perform safety checks and report falls. Consequently, the control methods for falling should expect challenging initial conditions where parts of the body are already tripping on the ground and falling is inevitable (Fig.\ref{subfig:tripping1}-\ref{subfig:tripping3}). To simulate the different initial conditions of the robot falling, we randomize the initial base and joint states and disable the joint actuators for a random period in each episode ranging from \unit[0.04]{s} to \unit[1.50]{s}. The upper time limit, \unit[1.50]{s} was chosen to allow late fall detection (robot falls completely, Fig.\ref{subfig:tripping5}). The control policy thus learns to react in such difficult situations and could be deployed on the robot as an emergency controller that the robot automatically switches to when the task controllers detect safety violations. After the initialization period, the policy controls the robot as elaborated in \ref{ssec:action}. The episodes only terminate at the end of the MDP{\color{Black}'s} time horizon.

\subsection{Asymmetric actor-critic} \label{ssec:aac}

We apply asymmetric actor-critic~\cite{pinto2017asymmetric} with Proximal Policy Optimization (PPO)~\cite{schulman2017proximal} such that the critic observes privileged observation features only available in the simulation. We define the observation vectors for the actor and the critic as the following:

\paragraph*{Actor observation} The actor observes the robot states, including the base orientation, base angular velocity, and joint states. The base linear velocity is excluded from the observation similar{\color{Black}ly} to~\cite{lee2019robust} because of the high uncertainty in the state estimation after the fall. Gaussian noise is added to the actor's observations during the training.

\paragraph*{Critic observation} {\color{Black}To improve the value estimation}, the critic observes noiseless actor observations and additional privileged observations {\color{Black}that are unavailable during the deployment but indicative of the expected return}. The privileged observations include the remaining time in the episode, the foot contact states, and other observations about the MDP, including a binary flag to indicate whether the actuators are activated and the time remaining in the initial phase of rollout until the actuators become active.

In particular, note that we only observe the remaining time to the end of the episode in the critic, and the actor's policy remains time-invariant. Maximizing the PPO objective function optimize{\color{Black}s} the original finite-horizon MDP. The privileged critic produces the value function estimate for this MDP, and the limited actor observation only constrains the structure of the policy. In this problem formulation, the algorithm trains towards the optimal time-invariant policy for the finite-horizon MDP.

\subsection{Actions} \label{ssec:action}

The joint target produced by the policy is computed as $(s \, a + \tilde{q})$, where $s$ is the action scaling factor, $a$ is the policy's action output, $\tilde{q}$ is the default joint position. The computed joint position is used as the position target for the drives' PD controller as suggested in \cite{peng2017SCAaction}. We do not use the joint difference action ({\color{Black}where} the position target is the sum of the scaled action and the current joint positions) because for ALMA, outputting a perturbation around the specified default angle (Fig.~\ref{subfig:default_joint_pos}) works well as a good initial policy. The default joint positions and the action scale are selected such that random actions have a chance to flip the robot when falling sideways. This helps explore the self-righting behavior in the policy's initial training iterations.

\subsection{Reward function} \label{ssec:rewards}

In this paper, we formulate fall damage reduction as minimizing a combination of the undesired measurements, such as high contact impulse and acceleration of bodies. In addition, the agent is rewarded at the end of the episode for standing up {\color{Black}in a configuration }close to ALMA's default stance pose. We define the fall and recovery problem as a finite{\color{Black}-}horizon MDP with time-based rewards similar to Rudin et al.~\cite{rudin2022advanced}, where time-variant task rewards are used to train efficient and adaptive locomotion skills on diverse terrains. The rewards that regularize the robot's undesirable behaviors, such as joint acceleration penalty and high impact, are time-invariant and active throughout the episode. For our problem, the rewards are categorized as {\color{Black} follows}:


\paragraph{Time-variant task rewards} The task reward for recovery contains three components. These three reward terms are activated only at the last \unit[2.0]{s} of each episode.

\textit{Base height:} The robot is rewarded for higher torso height. The maximum reward is obtained for height $\geq$ \unit[0.5]{m}.

\textit{Joint position:} Deviation from ALMA's default joint position is penalized.

\textit{Base orientation:} The robot is rewarded {\color{Black}for recovering} the base orientation {\color{Black} by penalizing the roll and the pitch angles}.

\paragraph{Time-invariant behavior rewards} {\color{Black}{To encourage smooth falling,}} behavior rewards penalizing body collision,  momentum change,  and body yank,  {\color{Black}{are used throughout the training episodes}}. The quantities required {\color{Black}for calculation of these terms, such as }the body acceleration and the contact forces, are accessed directly from the simulator.
\begin{table}[]
\caption{\textsc{Reward Terms Summary}}\label{tab:rewards}
\begin{tabular}{ccc}
Reward Term            & Definition                                                                                                                                                                                      & Scale                       \\ \hline  
\cellcolor{red!10}stand joint position &\cellcolor{red!10} $\begin{cases} e^{-\frac{\sum_j(q_j^*-q_j[t])^2}{\sigma_\text{p}N_j}} & \text{last \unit[2]{s}} \\ 0 & \text{otherwise} \end{cases}$ & 350  \\
\cellcolor{red!10}base height & \cellcolor{red!10}$\begin{cases} e^{-\frac{\max(h^*-h_\text{b}[t], 0))^2}{\sigma_\text{h}}} & \text{last \unit[2]{s}}\\ 0 & \text{otherwise} \end{cases}$ & 600 \\ 
\cellcolor{red!10}base orientation & \cellcolor{red!10}$\begin{cases} -g_\text{b} \cdot e_z & \text{last \unit[2]{s}}\\ 0 & \text{otherwise} \end{cases}$ & 120 \\
\cellcolor{blue!10} body collision        &          $\cellcolor{blue!10}\sum_{b\in \mathcal{B}} \lVert \lambda_\text{b}[t] \rVert^2$                                                                                                                                                                                       & -0.2                        \\
\cellcolor{blue!10} momentum change       &  \cellcolor{blue!10}                $\sum_{b\in \mathcal{B}} \lVert m_b a_b [t] \rVert$                                                                                                                                                                                 & $-5e^{-3}$ \\
\cellcolor{blue!10} body yank             &  \cellcolor{blue!10}             $\sum_{b\in \mathcal{B}} \lVert F_b[t]-F_b[t-1] \rVert^2 $                                                                                                                                                                                 & $-5e^{-2}$ \\
action rate           & $\sum (a[t] - a[t-1])^2$                                                                                                                                  & $-3e^{-3}$ \\
joint velocity        & $\sum_j \dot{q_j}^2 $                                                                                                                              & $-5e^{-4}$ \\
torques                & $\sum_j \tau_j ^2 $                                                                                                                                  & $-4e^{-7}$ \\
acceleration           & $\sum_j \ddot{q}_j^2$                                                                                                                              & $-1e^{-8}$
\end{tabular}
\end{table}

{\color{Black}{The reward terms and the scales are summarized in Tab.~\ref{tab:rewards}, where the task rewards are highlighted in red and the damage-reduction-related behavior rewards are in blue. In the reward definitions, $q_j$, $\dot{q}_j$, $\ddot{q}_j$ represent the angular position, velocity and acceleration of each joint. $\sigma_{(\cdot)}$ are the sensitivity scaling factor of the rewards, and $g_b$ is the projected gravity vector in the robot's base frame.}} $\mathcal{B}$ is the set of selected robot links, including the shanks, the thighs, the base, and all the arm links. $w$'s are the weights of the associated rewards. $m_b$, $a_b$, $\lambda_b$, and $F_b$ are the body mass, the body acceleration, contact force, and net force for body $b$ respectively.

Note that for the initial phase of the episode, when the robot's actuation is disabled, both the task and the behavior regularization rewards are set to zero since the policy does not affect the robot's joint actions during the period.

\subsection{Sim-to-Real}

%
We use NVIDIA {\color{Black}Isaac Gym}~\cite{makoviychuk2021isaac} to simulate the training environment. The simulator runs at 200~Hz, and we decimate the policy to run at 100~Hz. We implemented the following techniques to facilitate sim-to-real transfer.

\paragraph*{Actuator model}  An actuator model for the leg drives is used in the simulator similar to~\cite{hwangbo2019learning}. The pseudo-direct drives in the arm have better transparency than the serial elastic actuators in the legs. Therefore, we do not implement an actuator model but only randomize the friction and add torque delays in the arm drives in each training episode.

\paragraph*{Terrain randomization} {\color{Black}Uneven} terrain is used instead of {\color{Black}flat} terrain to randomize the ground contact normal direction and {\color{Black}encourage larger clearance when moving above the ground}.

\paragraph*{Observation noise} Gaussian noise is added to the actor observations during the training to make the policy robust against state estimation noises on the robot. 

\paragraph*{Robot randomization} The robot's base mass is randomized with $m_\text{rand}\sim \text{Uniform}(-5, 5)$ \unit[]{kg} additional mass sampled in each episode. The robot bodies' friction coefficients with the ground are also randomized.

%
%

\begin{figure*}
    \centering
    \subfloat[]{\includegraphics[width=0.156\textwidth]{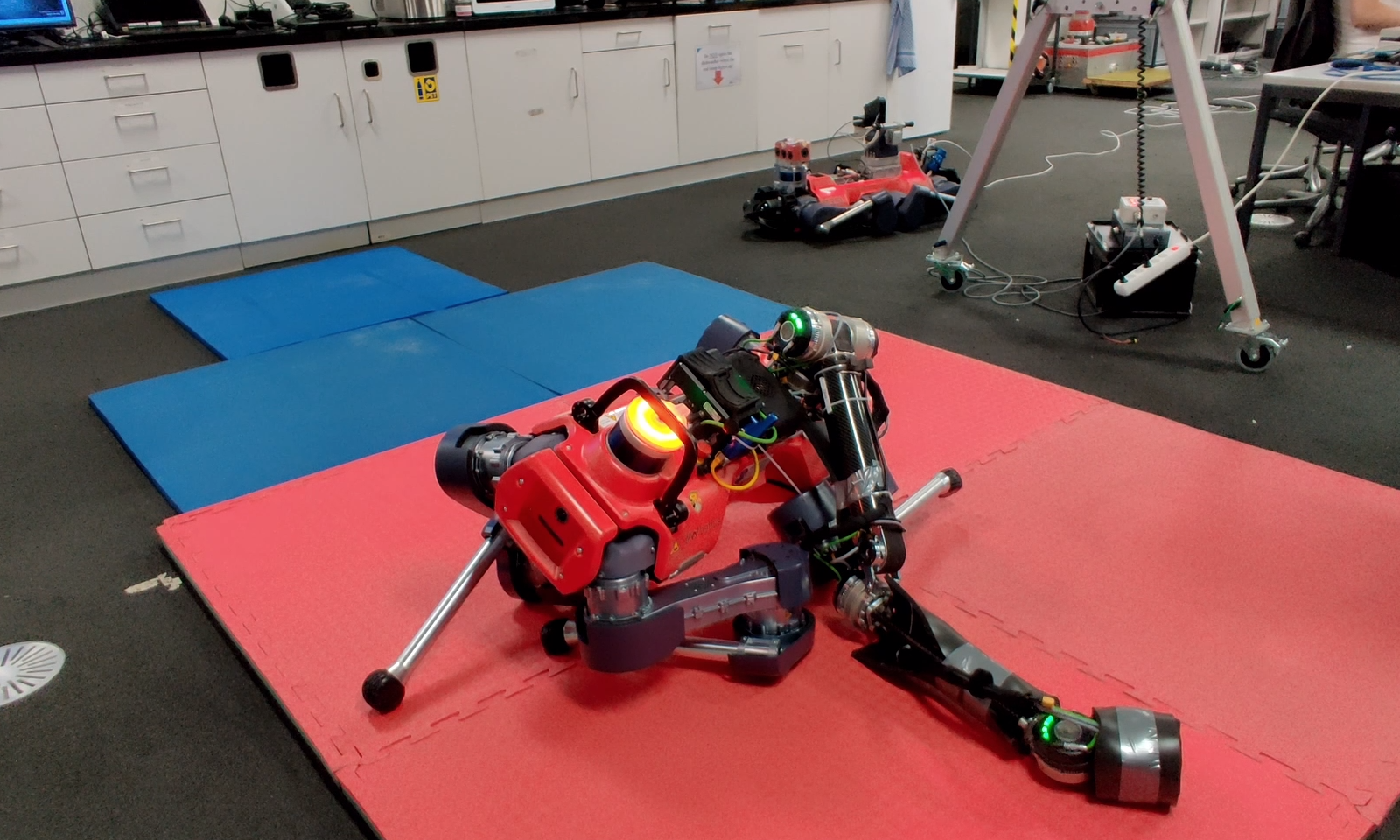}
    \vspace{-5pt}
    \label{subfig:recover_0}}
    \hfil
    \subfloat[]{\includegraphics[width=0.156\textwidth]{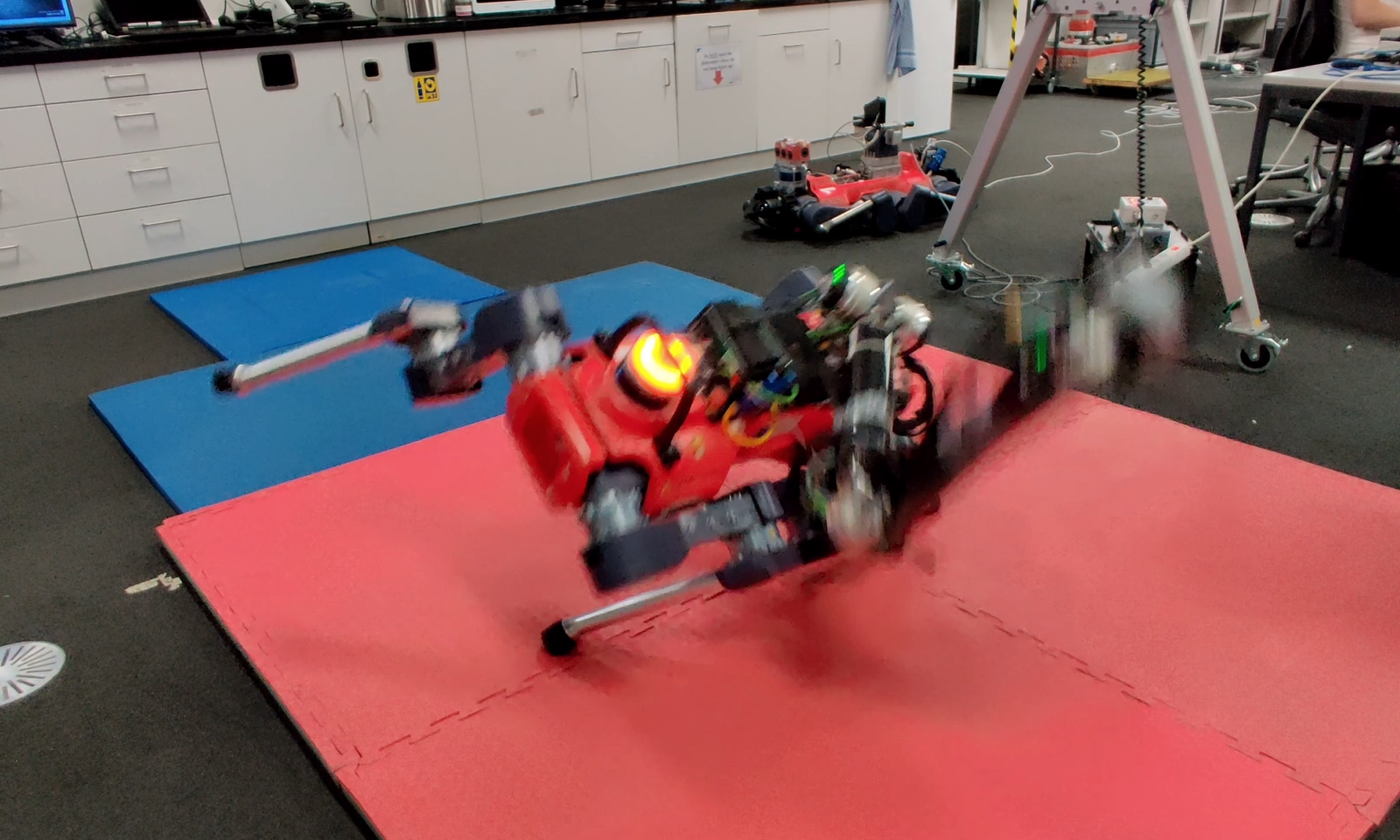}
    \vspace{-5pt}
    \label{subfig:recover_1}}
    \hfil
    \subfloat[]{\includegraphics[width=0.156\textwidth]{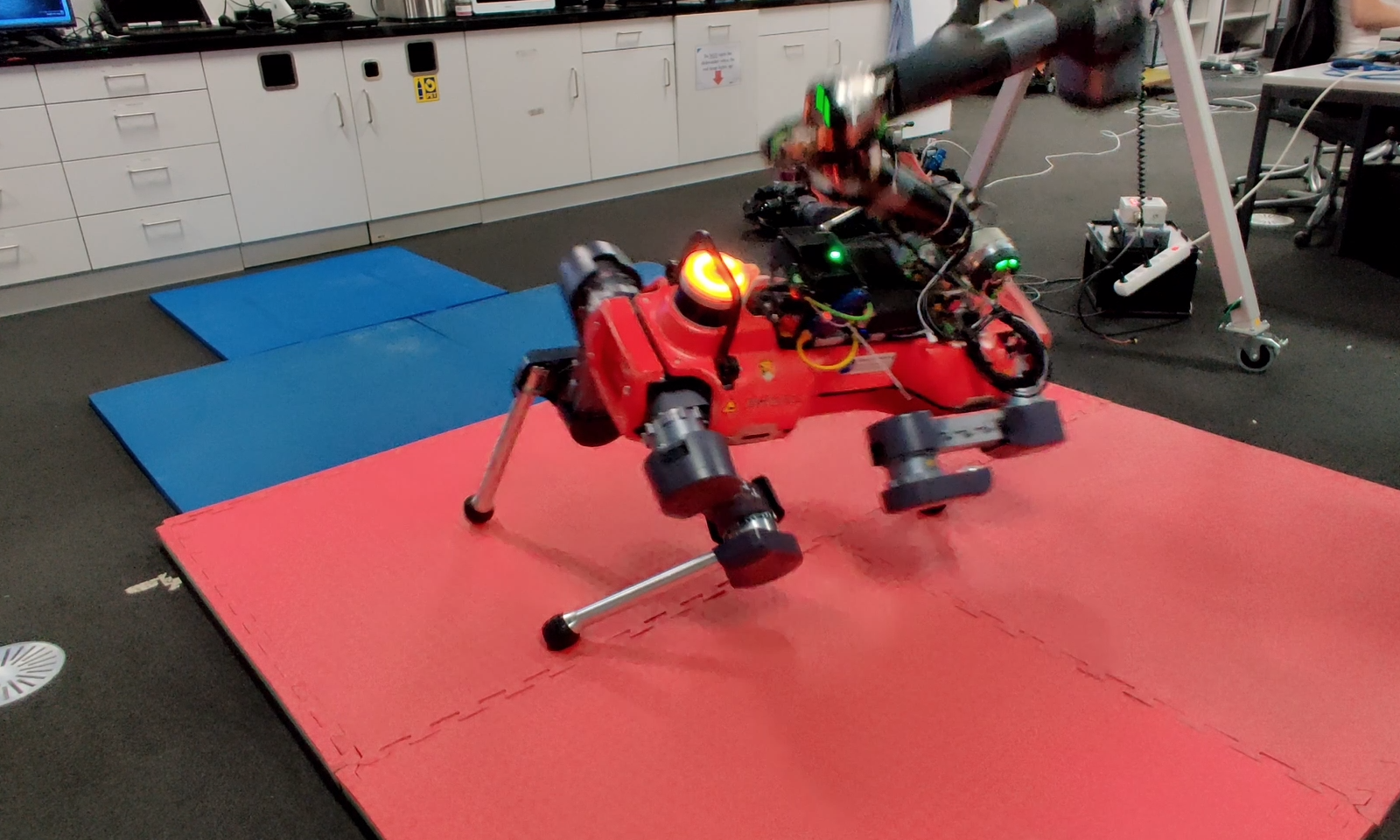}
    \vspace{-5pt}
    \label{subfig:recover_2}}
    \hfil
    \subfloat[]{\includegraphics[width=0.156\textwidth]{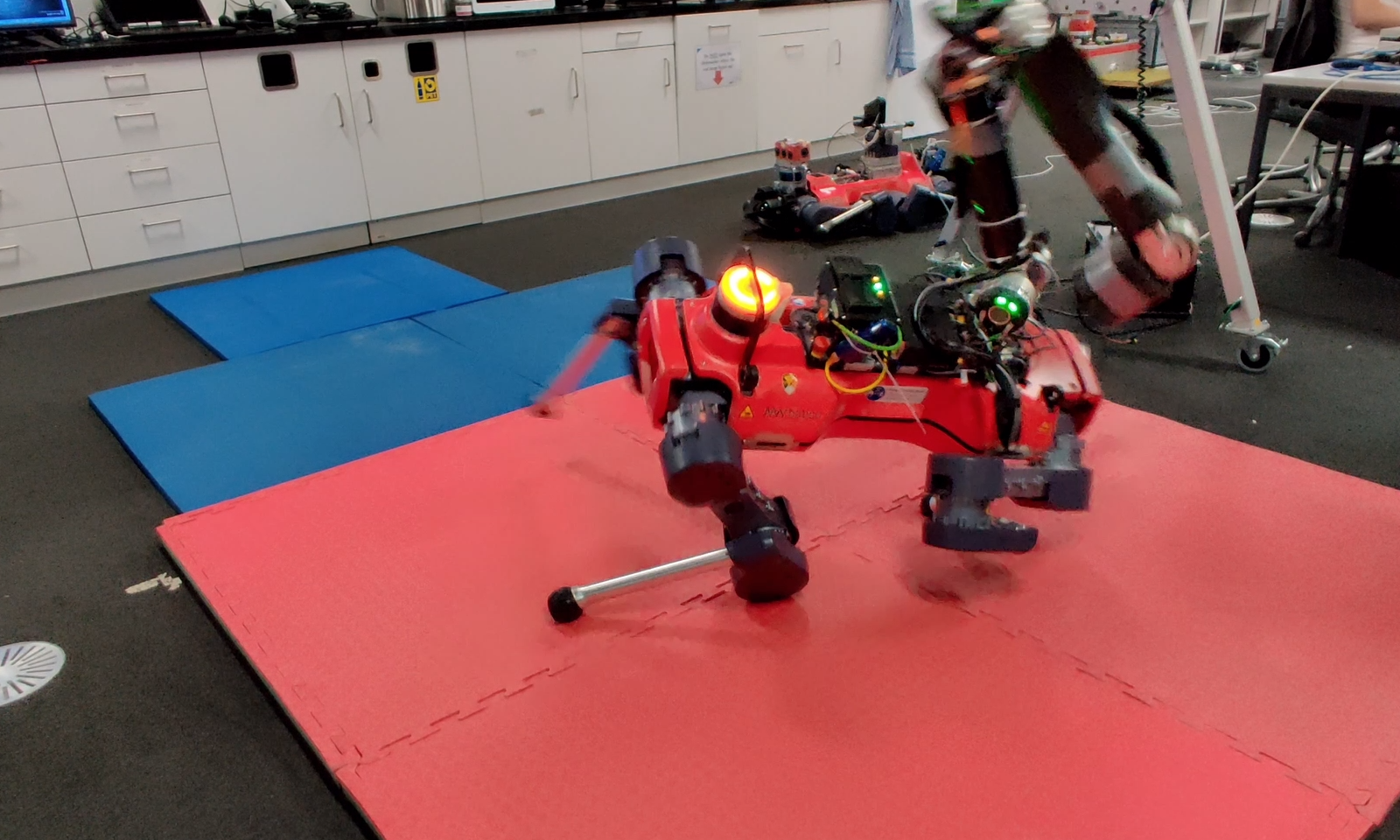}
    \vspace{-5pt}
    \label{subfig:recover_3}}
    \hfil
    \subfloat[]{\includegraphics[width=0.156\textwidth]{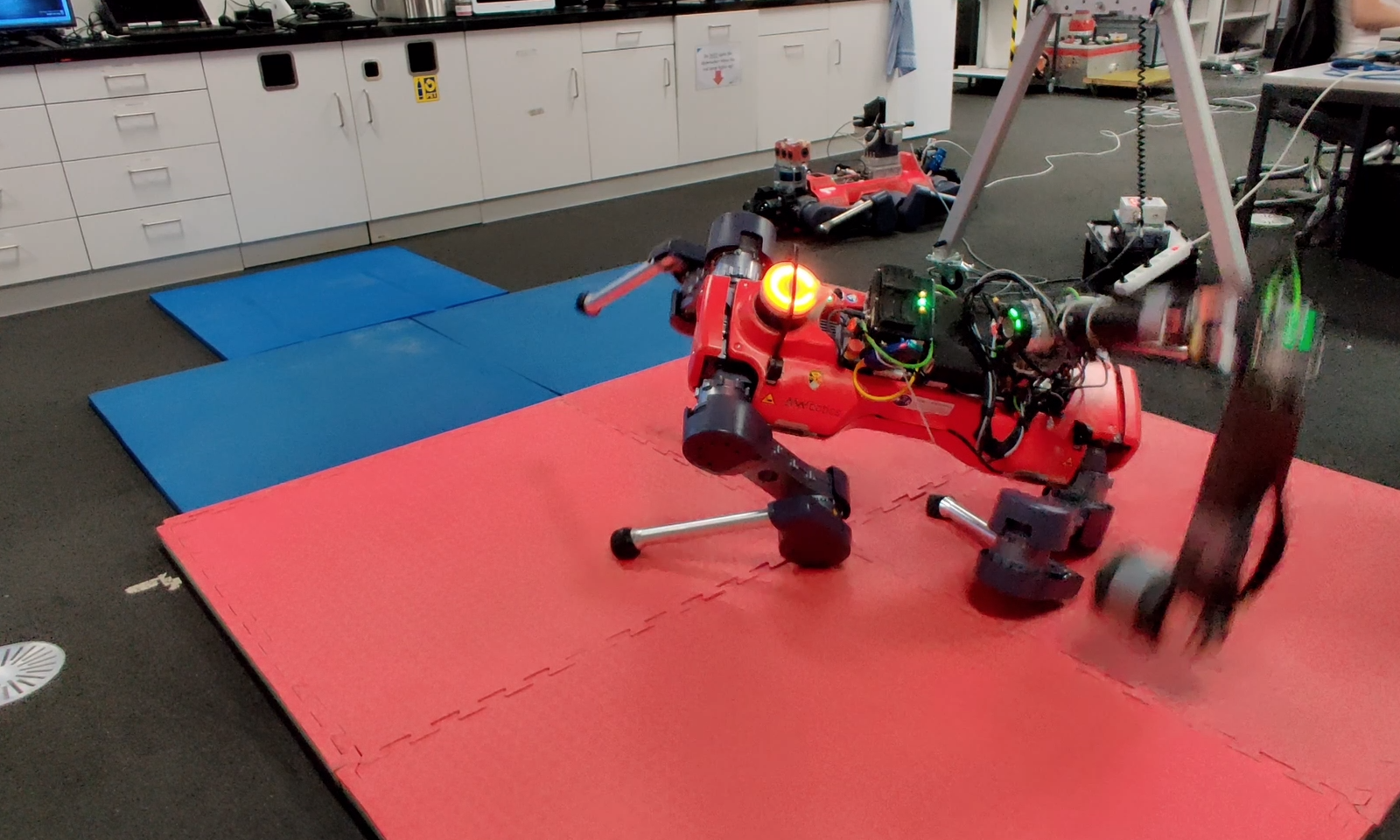}
    \vspace{-5pt}
    \label{subfig:recover_4}}
    \hfil
    \subfloat[]{\includegraphics[width=0.156\textwidth]{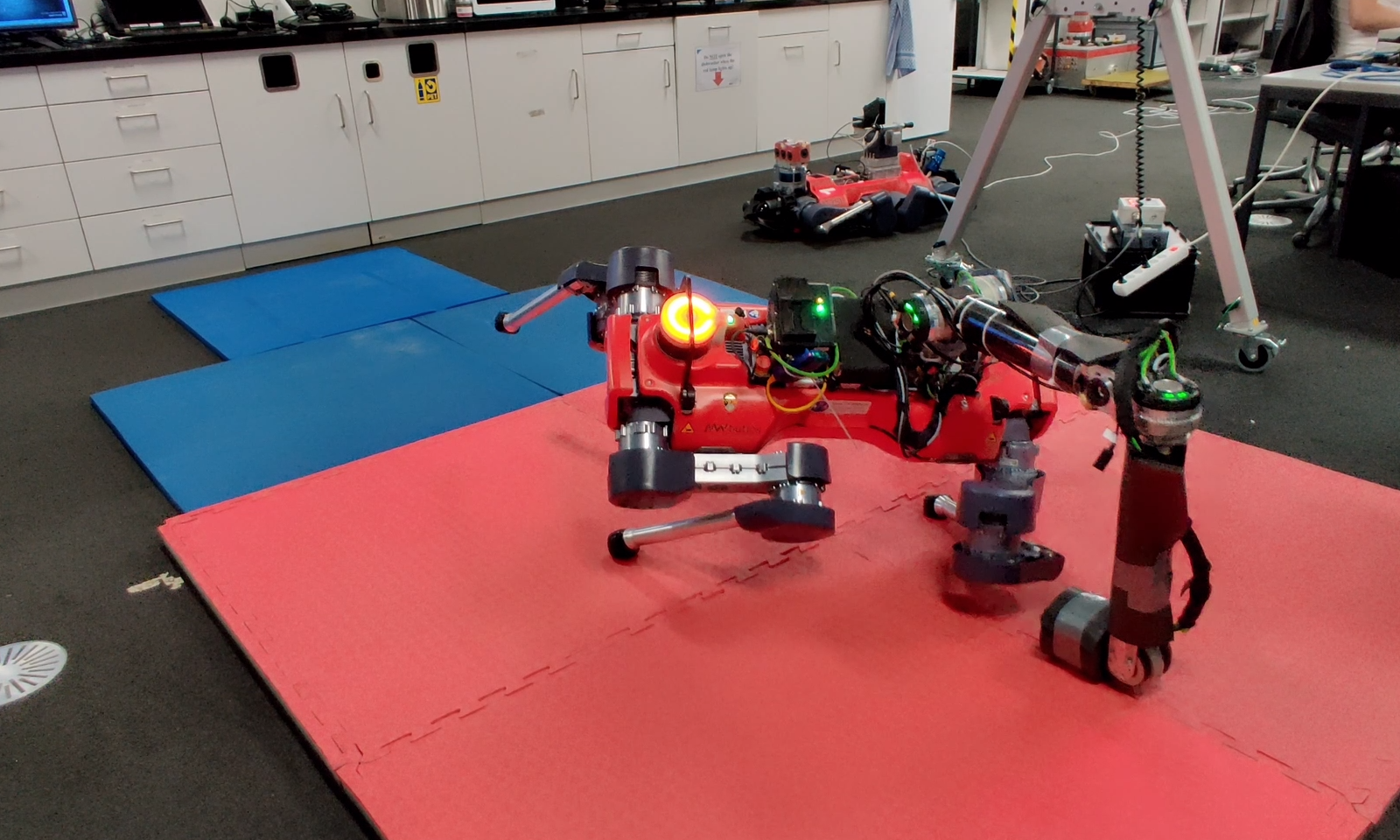}
    \vspace{-5pt}
    \label{subfig:recover_5}}
    \hfil
    \subfloat[]{\includegraphics[width=0.156\textwidth]{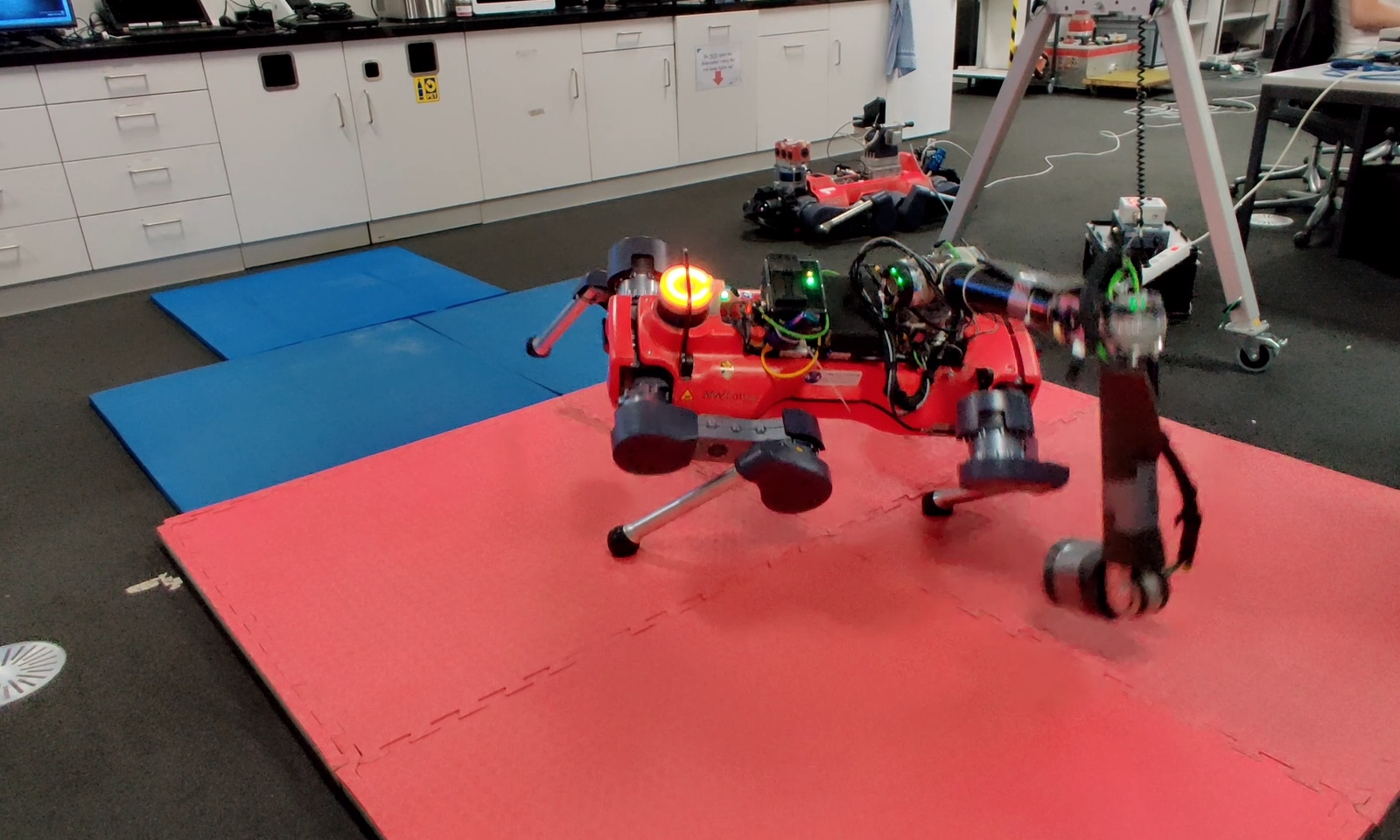}
    \vspace{-5pt}
    \label{subfig:recover_6}}
    \hfil
    \subfloat[]{\includegraphics[width=0.156\textwidth]{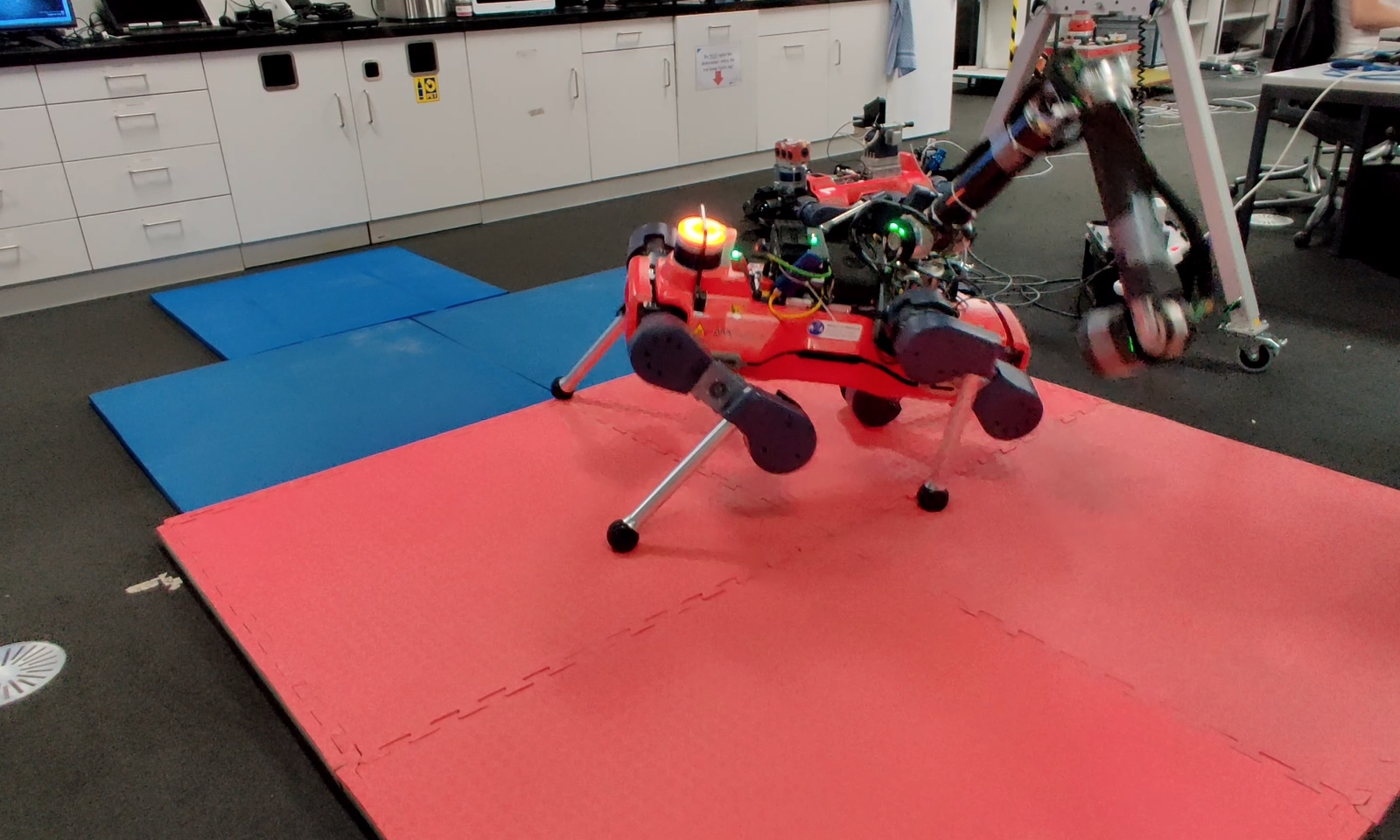}
    \vspace{-5pt}
    \label{subfig:recover_7}}
    \hfil
    \subfloat[]{\includegraphics[width=0.156\textwidth]{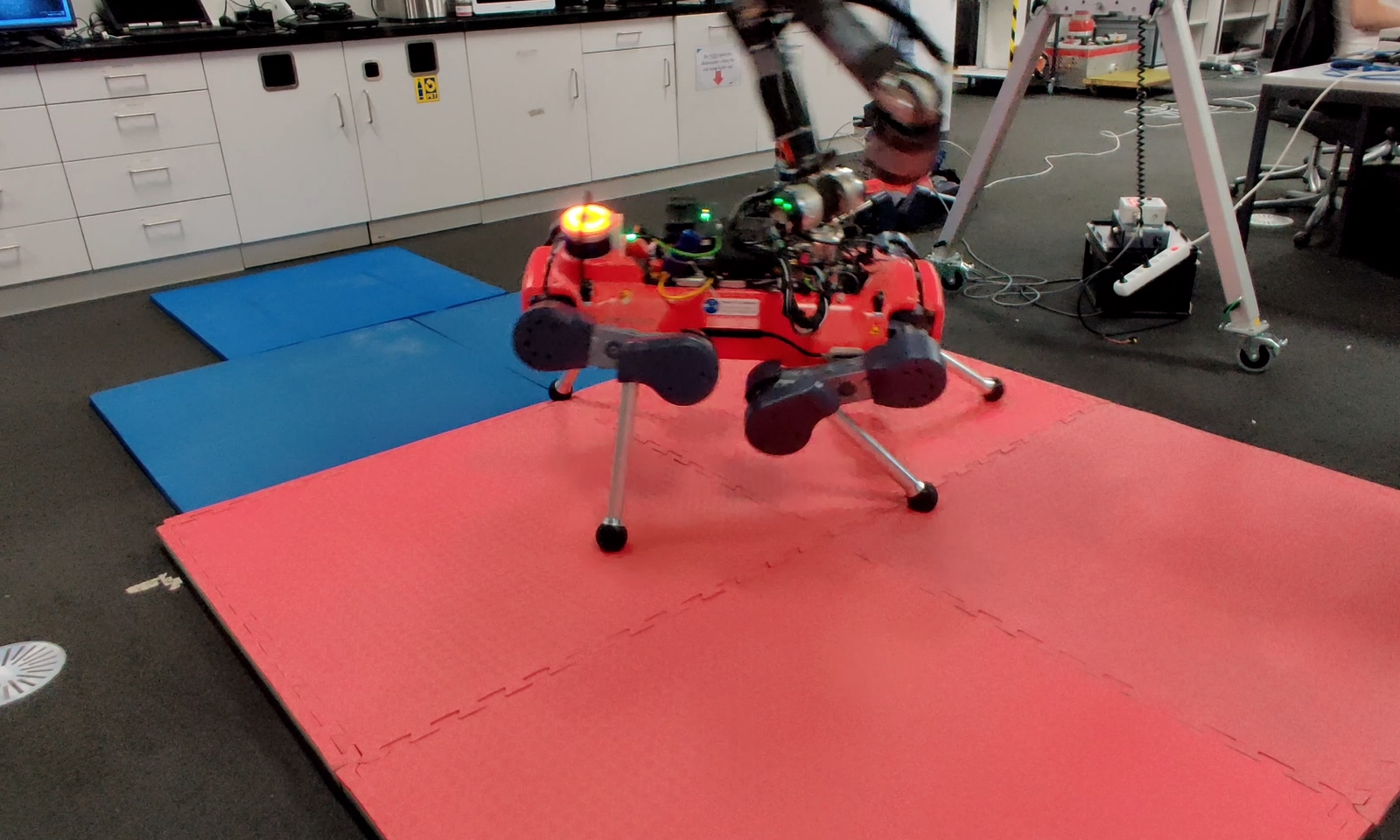}
    \vspace{-5pt}
    \label{subfig:recover_8}}
    \hfil
    \subfloat[]{\includegraphics[width=0.156\textwidth]{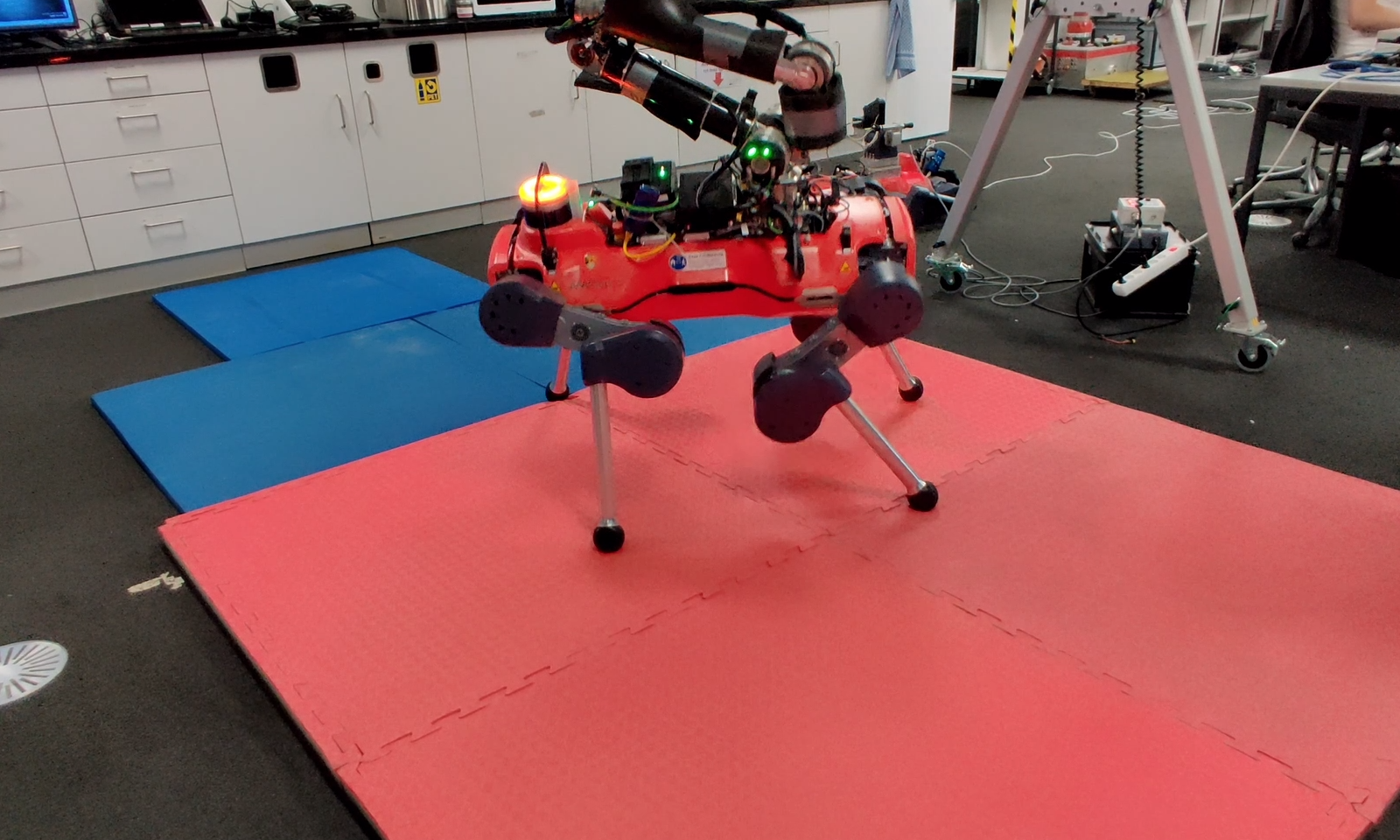}
    \vspace{-5pt}
    \label{subfig:recover_9}}
    \hfil
    \subfloat[]{\includegraphics[width=0.156\textwidth]{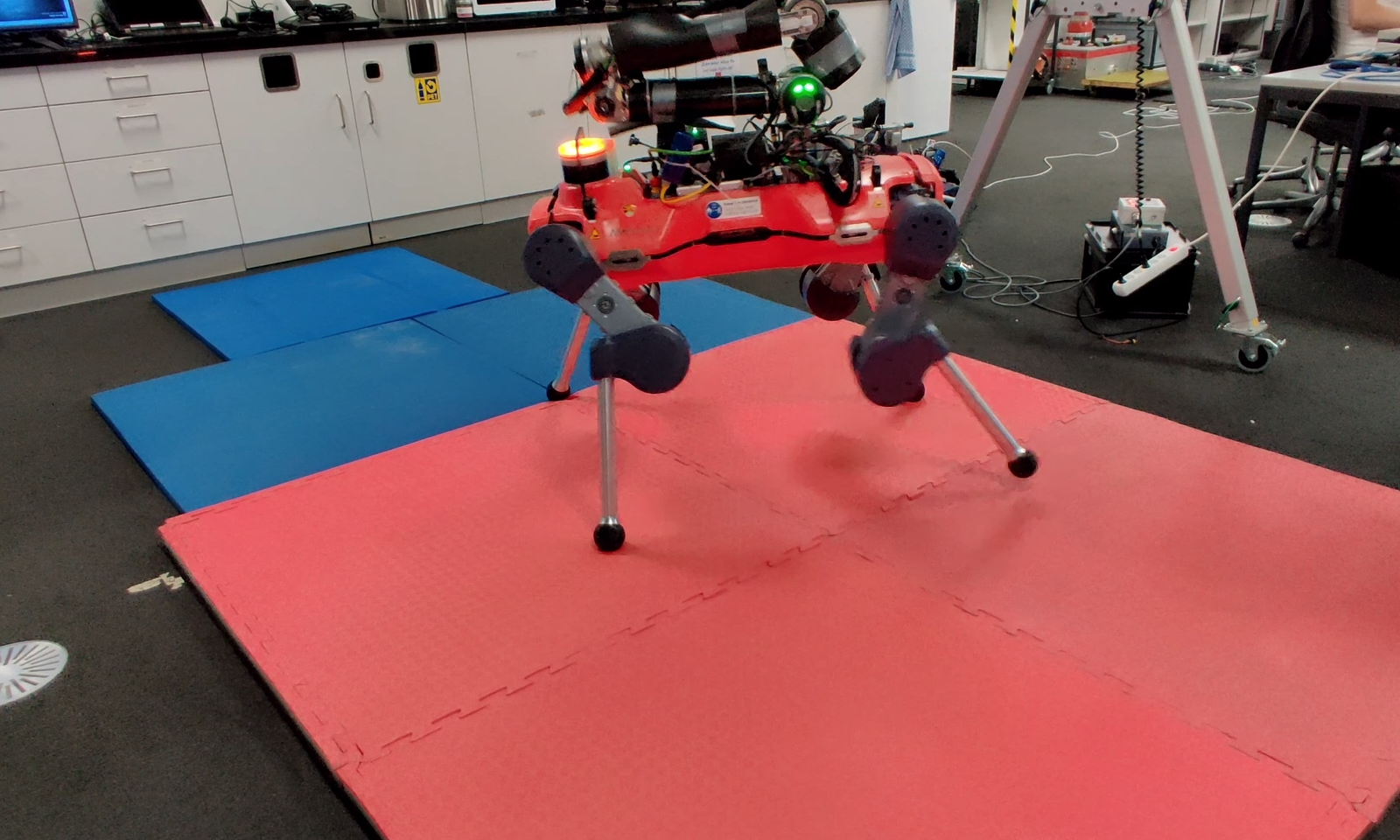}
    \vspace{-5pt}
    \label{subfig:recover_10}}
    \hfil
    \subfloat[]{\includegraphics[width=0.156\textwidth]{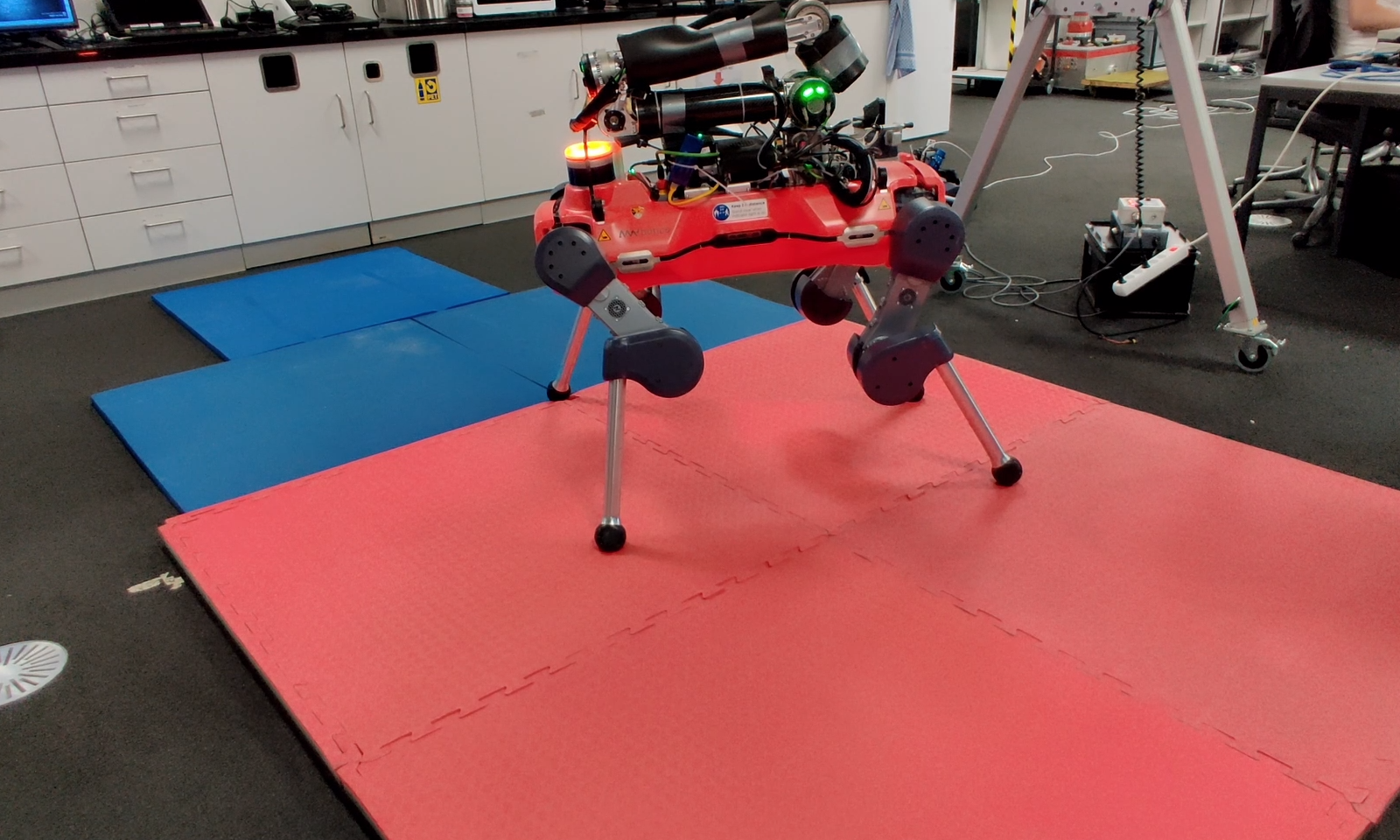}
    \vspace{-5pt}
    \label{subfig:recover_11}}
    \hfil
    \vspace{5pt}
    \caption{ALMA robot adapting the fall recovery strategy after an unsuccessful quick recovery attempt.}
    \vspace{-10pt}
    \label{fig:recover_main}
\end{figure*}

\section{Results}

\subsection{Fall damage reduction}


We evaluate the policy's performance {\color{Black}in} falling by comparing the peak instantaneous impulse on the base, mean and peak base acceleration, and peak joint internal forces across all joints. This follows the evaluation criteria in \cite{ha2015multiple,kumar2017learning,ruiz2009learning}. The max peak joint internal forces are used as one of the evaluation criteria because the high stress induced on the drives during the impact is a common cause of damage for heavy robots. The evaluation is carried out with 2560 test episode rollouts with randomized initial fall configuration and base mass, and the performance is compared to two baseline emergency controllers {\color{Black}{with the originally deployed settings}}, namely freezing and damping the drives. 

\begin{figure}
    \centering
    \subfloat[]{\includegraphics[width=0.245\textwidth]{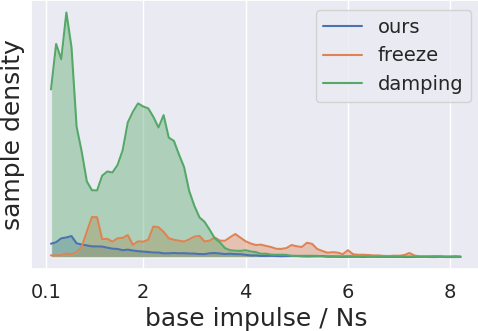}
    \vspace{-3pt}
    \label{subfig:base_contact_force}}
    \hfil
    \subfloat[]{\includegraphics[width=0.22\textwidth]{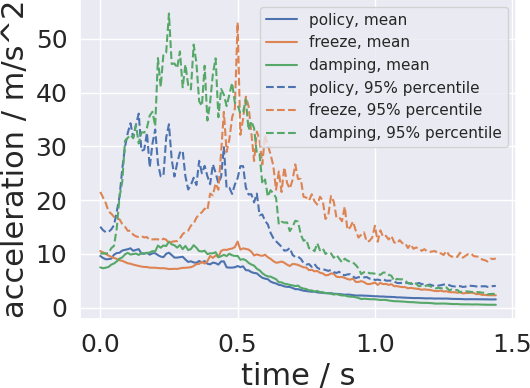}
    \label{subfig:base_acc}}
    \hfil
    \vspace{1pt}
    \caption{Left: Distribution of contact impulse on the base over all time steps. Time steps with base contact impulse below \unit[0.05]{Ns} are not included. Right: Base acceleration during the fall.}
    \vspace{-1pt}
    \label{fig:base_contact}
\end{figure}

Fig.~\ref{subfig:base_contact_force} shows the distribution of the base contact impulse across all time steps in all test runs. Time steps with base contact impulse below \unit[0.05]{Ns} are not included in the plot because this load is not expected to damage the robot base. The low{\color{Black}{er}} number of samples for {\color{Black}{our proposed}} policy in this plot indicates that the policy avoids contact impulses above \unit[0.05]{Ns} when the robot falls. In contrast, the damping controller exhibits two peaks, corresponding to the base impulse when falling directly down and sideways, respectively. Freezing the drive leads to a flat tail in the distribution, indicating a non-negligible probability of a high base impulse during the fall. The reduction in the base damage is also illustrated in Fig.~\ref{subfig:base_acc}, where the mean and 95th percentile of base acceleration at each time step during the rollout is compared. This additionally compares the proposed method with the baseline controllers in the worst-case scenario as it is less visible in the impulse distribution plot. The 95th percentile base acceleration is significantly lower for our learning-based policy than the baseline controllers. 

\begin{figure}
    \centering
    \subfloat[]{\includegraphics[width=0.22\textwidth]{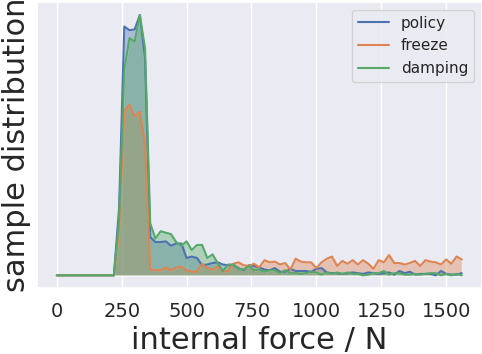}
    \vspace{-5pt}
    \label{subfig:peak_int}}
    \hfil
    \subfloat[]{\includegraphics[width=0.22\textwidth]{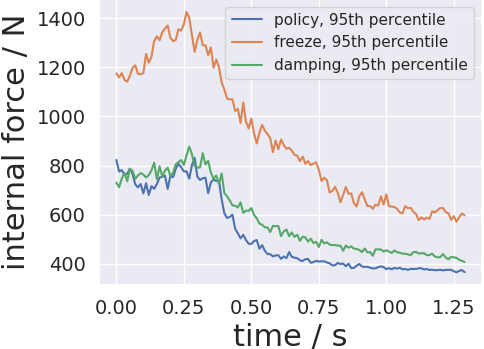}
    \vspace{-5pt}
    \label{subfig:int_dist}}
    \hfil
    \vspace{1pt}
    \caption{Left: Distribution of peak joint internal force on the base over all time steps. Right: 95th percentile of joint internal force during the fall.}
    \vspace{-4pt}
    \label{fig:internal_force}
\end{figure}

Fig.~\ref{subfig:peak_int} and \ref{subfig:int_dist} indicate a marginal improvement in the distribution of the peak joint internal force for the proposed control policy over the damping controller {\color{Black}{and our fall-recovery controller}. In both cases, we observe a significant decrease in peak internal force than the freezing strategy}. Overall, our proposed method reduces the base contact impulse and the acceleration without incurring higher peak internal joint forces.


\subsection{Fall recovery}

\begin{figure*}
    \centering
    \includegraphics[width=0.98\textwidth]{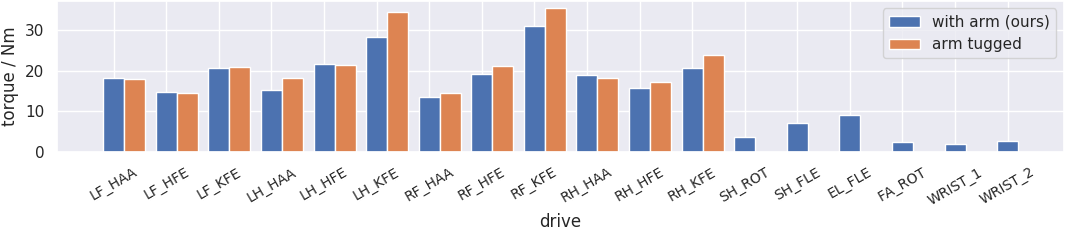}
    \hfil
    \vspace{5pt}
    \caption{The mean drive torques of the policy with and without using the arm. Both policies converged to a non-symmetric strategy of using higher torques in the Left-Hind and Right-Front legs. However, the mean torques for these two legs are lower for the policy that uses the arm.}
    \vspace{-9pt}
    \label{fig:arm_vs_no_arm}
\end{figure*}


The policy learns to use the arm adaptively for falls and recovery. Fig.~\ref{fig:recover_main} shows an example of the adaptive behavior when recovering from a fall. In this example, the policy first attempted a quick recovery by jumping up with the right legs and balancing by extending the left legs (\ref{subfig:recover_1}). {\color{Black}However, t}his attempt was unsuccessful, and the robot lost balance (\ref{subfig:recover_2}). The robot {\color{Black}then }adapted {\color{Black}its} strategy and used the arm and the right-front knee to secure the fall (\ref{subfig:recover_3}-\ref{subfig:recover_4}). From this configuration, the robot pushed {\color{Black}against} the ground with {\color{Black}its} arm to recover (\ref{subfig:recover_5}-\ref{subfig:recover_6}) and subsequently retracted the arm and adjusted the stance (\ref{subfig:recover_7}-\ref{subfig:recover_11}). A different recovery strategy is also shown in Fig.\ref{fig:title}.







The arm-assisted recovery motion is compared {\color{Black}to} a {\color{Black}baseline learning-based controller}, where the arm is frozen and tugged throughout the episode {\color{Black}and the policy only uses the legs for recovery}. Both controllers are trained with the same reward and MDP setup except for the arm. We label a fall recovery maneuver successful if, at the end of the episode, the base height is above \unit[0.5]{m} and the max joint velocity is below \unit[0.01]{rad/s}. 2560 episodes were tested with each policy for comparison.
We observe that the arm-assisted recovery policy achieves a 98.9\% success rate compared to 95.2\% {\color{Black}{with}} the tugged-arm policy. Fig.~\ref{fig:arm_vs_no_arm} notes the torque consumption in each drive for the falling and recovery scenarios averaged over the time steps and the test episodes.  Without help from the arm, the robot needs to tug its legs further to push and flip itself when falling sideways. We observe using the arm for recovery reduces the average leg torque consumption by 9.17\% {\color{Black}averaged} over all the leg drives. 


\subsection{Ablation for the observation configurations}

\begin{table}[t]
\caption{\textsc{Comparing the Actor and Critic Observation Configurations}} \label{tab:aac}
\begin{tabular}{ll|ll}
actor obs               & critic obs              & episode return & value error \\ \hline
1.($o_\text{s}$)                  & ($o_\text{s}$)                   & -3.88            & 0.0902    \\
2.($o_\text{s}$) \textbf{(ours)}                 & ($o_\text{s}$, $o_\text{priv}$, $o_\text{MDP}$) & 12.9            & 0.00379   \\ 
3.($o_\text{s}$, $o_\text{eplen}$)        & ($o_\text{s}$, $o_\text{priv}$, $o_\text{MDP}$) & 12.9            & 0.00411   \\ 
4.($o_\text{s}$, $o_\text{priv}$, $o_\text{MDP}$) & ($o_\text{s}$, $o_\text{priv}$, $o_\text{MDP}$) & 13.3          & 0.00336  \\ \hline
\end{tabular}

\begin{align*}
o_\text{s} &\text{: state observations}\\
o_\text{priv} &\text{: privileged observations (contact, noiseless state estimation)}\\
o_\text{eplen} &\text{: current episode length observations}\\
o_\text{MDP} &\text{: MDP observations (episode length, initialization flag)}
\end{align*}

\end{table}

We compare different observation configurations for the asymmetric actor-critic setup. The configurations are summarized in Tab.~\ref{tab:aac}, where the evaluation values are the mean of three {\color{Black}random} seeds after training for 20000 iterations (equivalent to $3.93\times10^9$ environment steps). We evaluate the performance of the policies based on the episode return and the value error. The value error is the mean squared error between the value estimation from the critic and the discounted episode return. {\color{Black}This metric} indicates how accurately the critic provides the baseline function for the PPO policy update. We compare the following {\color{Black}combinations} of actor and critic observations.

\paragraph{Privileged critic}

In the first configuration, the policy is trained with a critic that does not observe {\color{Black}{the episode progress}} or privileged robot state observations. It can be seen that this configuration is unable to learn the task. {\color{Black}{We speculate that this is because the non-privileged critic is time-invariant and thus unable to evaluate the {\color{Black}true} time-variant value function {\color{Black}induced by} time-{\color{Black}varying} reward {\color{Black}function}s, leading to significantly higher variance in the policy update.}} In contrast, the critic function that has access to privileged observation has a drastically lower value function estimation error, allowing the non-privileged actor to learn fall damage reduction and recovery skills.

\paragraph{Time-variant vs. time-invariant actor}

\begin{figure}
    \centering
    \subfloat[]{\includegraphics[width=0.15\textwidth]{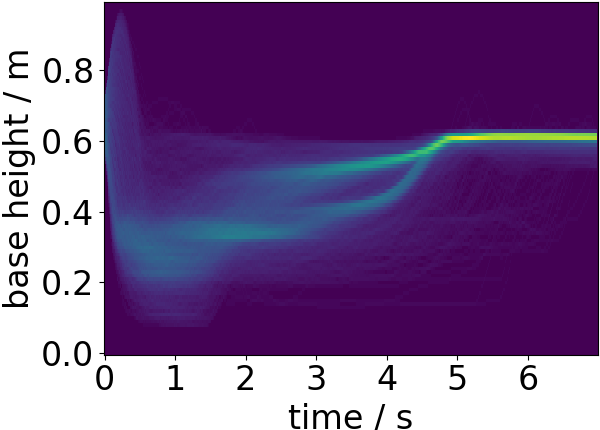}
    \vspace{-5pt}
    \label{subfig:sym_height}}
    \hfil
    \subfloat[]{\includegraphics[width=0.15\textwidth]{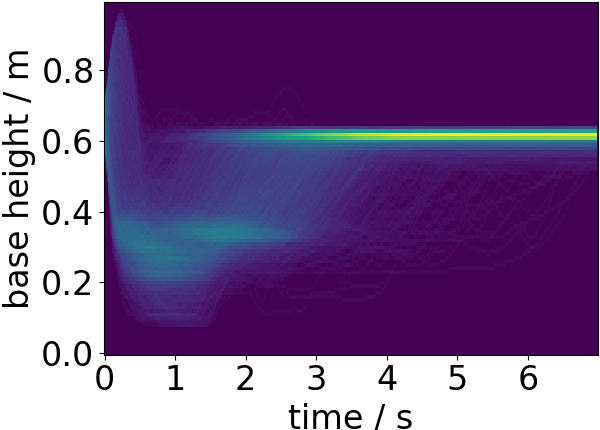}
    \vspace{-5pt}
    \label{subfig:asym_height}}
    \hfil
    \subfloat[]{\includegraphics[width=0.15\textwidth]{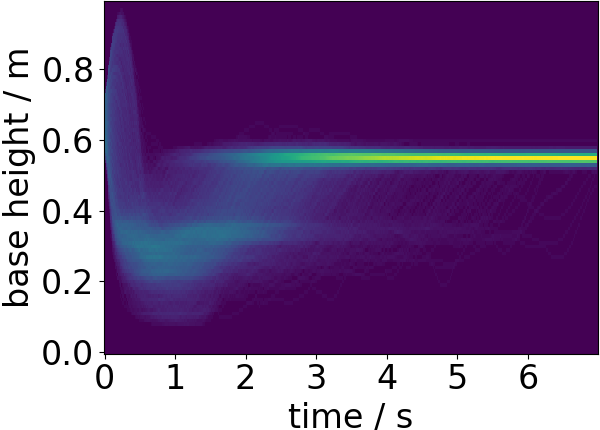}
    \vspace{-5pt}
    \label{subfig:large_rew_height}}
    \hfil
    \vspace{5pt}
    \caption{The recovery {\color{Black}{profile}} (indicated with the distribution of base height) of the time-variant {\color{Black}{(\ref{subfig:sym_height})}} and time-invariant policy {\color{Black}{(\ref{subfig:asym_height})}} for the finite-horizon MDP. While both policies complete fall recovery within the specified time horizon, the time-invariant policy shows a more uniform behavior. {\color{Black}{Fig.~\ref{subfig:large_rew_height} shows the recovery profile with the task rewards increased by three times. Its similarity to Fig.~\ref{subfig:asym_height} indicates the robustness of the time-based reward scheme against reward scaling.}}}
    \vspace{-1pt}
    \label{fig:asymmetric}
\end{figure}

Comparing configurations 2 and 3, including the remaining time of the episode in the actor's observation does not result in a significant change in the policies' episode returns, but it leads to different recovery behaviors of the actors. Fig.~\ref{fig:asymmetric} plots the distribution of the base height over the test episodes against time and illustrates the effect of this observation on the recovery behavior. The pixels with a lighter color in the plot indicate a higher number of episodes with the corresponding height at the timestep. The time-variant policy often produces the behavior such that the robot stands halfway and rests on the joint limits of the Hip Abduction-Abduction drives. It then rapidly reorganizes its legs to the default standing configuration shortly before the task rewards are activated. We think this behavior is learned because it reduces the torque consumption in the behavior reward without lowering the task rewards. However, this is undesirable in practice because the motion is not uniform in time. Additionally, in case of a failed self-righting attempt, the time-variant policy no longer tries to recover if the expected motions penalty is higher than the expected return in the remaining time. In contrast, the time-invariant policy exhibits a temporally consistent behavior and always attempts to recover. Hence our proposed configuration is more suitable and resilient for deployment. 

\paragraph{Asymmetric actor-critic vs. privileged policy}

Our proposed configuration with non-privileged, time-invariant actor and privileged critic (configuration 2) results in a 3.0\% decrease in the mean episode return compared to the privileged policy setup (configuration 4). However, the latter setup requires privileged actor observation and thus cannot be directly deployed on the robot. For the privileged policy, additional training steps need to be implemented to distill another policy that only uses non-privileged observation for hardware deployment as proposed in \cite{lee2020learning}.

\subsection{{\color{Black}{Reproducibility and adaptation}} to other tasks}
{\color{Black}{We verify the robustness of the proposed method to reward scaling by comparing the recovery profiles of two policies: one trained with the original reward scale (Fig.~\ref{subfig:asym_height}) and another trained with three times the original time-variant task rewards (Fig.~\ref{subfig:large_rew_height}). Both policies learn to recover before the activation of time-variant task rewards and optimize behavior rewards, leading to similar height distributions over the episode.}}

We note that the training pipeline can {\color{Black}{be}} trivially {\color{Black}{modified for}} other state-transition tasks for legged mobile manipulators by only modifying high-level settings such as the rollout initialization period, the total task duration, the joint target position, and the reward scales. In the following, we outline the changes required to {\color{Black}{train}} policies to perform two maneuvers: resting on the ground from any stance configuration and self-righting from a fallen state to the default resting position on the floor. The two policies are also validated on hardware (shown in the linked video).

\paragraph{Resting} In the initialization period, the drives are not disabled, and the policy is rewarded for standing at a target height with randomly sampled desired joint positions. This step is required to randomize the initial stance configuration before executing the resting motion because it is hard to directly reset the robot to random stance poses on uneven terrain in the simulator. The behavior reward scales prioritize reducing the joint velocity over the torque consumption in this task.

\paragraph{Self-righting} The initialization period length is set to \unit[2.0]{s} so that the robot always falls completely. The target joint configuration for this policy is the robot's default resting configuration.

For both tasks, our training pipeline discovers natural contact sequences that reduce the peak impact on the base while going to the desired final joint configuration at the end of each task. 
%
%
\section{Conclusion}
This work presents an asymmetric actor-critic method to train a time-invariant control policy for legged mobile manipulators. The policy is trained in simulation under randomized fall configurations with time-based rewards. It learns to use the arm for falling {\color{Black}damage reduction} and recovery adaptively. For fall damage reduction, the proposed controller improves over the baselines emergency controllers in peak instantaneous impulse, base acceleration, and peak joint internal forces during the fall. It also outperforms the {\color{Black}learning-based }arm-tugged recovery {\color{Black}policy} in both recovery success rate and leg torque consumption. We extensively tested and deployed the policy in simulation and on hardware for ALMA robot.

One of the shortcomings of the current implementation is that the training environment avoids the base contacting the ground but does not {\color{Black}adapt to potential damages such as dysfunctional drives due to the fall. We plan to extend the policy to consider these cases in the future by exploring policy architectures that allow for quick adaptation for such scenarios.}

\section*{Acknowledgements}

We would like to thank Joonho Lee, Nikita Rudin, Firas Abi-Farraj, and Jia-Ruei Chiu for their suggestions on the training environment. We also thank Fan Yang, Eris Sako, and Jan Preisig for their help with the experiments.

%

\clearpage
\bibliographystyle{IEEEtran}
\bibliography{sources} 
\end{document}